# Continuous Bangla Sign Language Translation: Mitigating the Expense of Gloss Annotation with the Assistance of Graph

Rabeya Akter
Roll: 139303

Safaeid Hossain Arib
Roll: 139311

Supervisor:
Dr. Sejuti Rahman
Associate Professor

A Project submitted for the degree of
B.Sc. Engineering in Robotics and Mechatronics Engineering

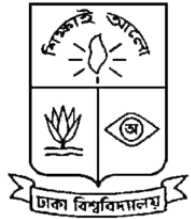

Department of Robotics and Mechatronics Engineering
University of Dhaka, Dhaka-1000, Bangladesh

August 2025

# Abstract

Millions of individuals worldwide are affected by deafness and hearing impairment. Sign language serves as a sophisticated means of communication for the deaf and hard of hearing. However, in societies that prioritize spoken languages, sign language often faces underestimation, leading to communication barriers and social exclusion. The Continuous Bangla Sign Language Translation project aims to address this gap by enhancing translation methods. While recent approaches leverage transformer architecture for state-of-the-art results, our method integrates graph-based methods with the transformer architecture. This fusion, combining transformer and STGCN-LSTM architectures, proves more effective in gloss-free translation. Our contributions include architectural fusion, exploring various fusion strategies, and achieving a new state-of-the-art performance on diverse sign language datasets, namely RWTH-PHOENIX-2014T, CSL-Daily, How2Sign, and BornilDB v1.0. Our approach demonstrates superior performance compared to current translation outcomes across all datasets, showcasing notable improvements of BLEU-4 scores of 4.01, 2.07, and 0.5, surpassing those of GASLT, GASLT and slt_how2sign in RWTH-PHOENIX-2014T, CSL-Daily, and How2Sign, respectively. Also, we introduce benchmarking on the BornilDB v1.0 dataset for the first time. Our method sets a benchmark for future research, emphasizing the importance of gloss-free translation to improve communication accessibility for the deaf and hard of hearing.

# Contents

# List of Figures

# List of Tables

# Chapter 1

# Introduction

## 1.1   Deafness and Hearing Loss

Millions of individuals throughout the world are afflicted with deafness and hearing loss. Nearly 20 percent of the world's population, or more than 1.5 billion individuals, suffer from hearing loss. [15] partial or complete inability to perceive sounds in either one or both ears is refered to as hearing loss, whereas complete or partial loss of hearing is deafness. All ages of people can be affected by these diseases, which can be brought on because of a number of things including heredity, infections, trauma, and exposure to loud noises. More than 700 million people may have a possibility of experiencing significant hearing impairment by the year 2050 [15]

Living with hearing loss or deafness can be difficult since it can limit a person's capacity for social interaction, communication, and engagement with their environment. However, persons with these disorders can frequently improve their hearing and lead full lives with the aid of contemporary technology and treatments.

Deafness and hearing loss should be made more widely known because early detection and treatment can significantly improve a person's quality of life. Furthermore, by being aware of the difficulties faced by people with hearing loss, we can build a society that is more tolerant and inclusive.

## 1.2    Sign Languages

Individuals who are deaf or hard of hearing require a means to interact and communicate with others. They do this by using sign language. Around 70 million individuals with hearing challenges, including both the hearing-impaired and deaf individuals worldwide, utilize sign language in order to communicate.[16] A unique and complex form of communication, sign language uses body language, face expressions, and hand gestures to transmit messages. Deaf or hard of hearing people who struggle with spoken language employ this visual language. In contrast to spoken languages, sign languages are not universal and there are many different sign languages spoken around the world, each having its own grammar, syntax, and vocabulary. All over the globe, there are more than 300 sign languages used for communication. [17]Numerous sign languages are there, including Bangla Sign Language (BdSL), American Sign Language (ASL), Chinese Sign Language (CSL). Despite its importance as the primary form of communication for millions of people worldwide, sign language is frequently ignored in cultures where spoken language is valued more highly. For those who rely on sign language, this negligence can lead to communication issues, marginalization, and discrimination. Therefore, it is crucial to acknowledge the worth and legitimacy of sign language while considering the special and essential nature and to work towards creating more inclusive and accessible communities that make it easier for everyone who needs it to utilize sign language.

## 1.3    Sign Language Recognition(SLR)

Building technology that can automatically read and translate sign language into text or voice is the aim of research in the newly growing field of recognition of sign language. Communication barriers between people who utilize spoken language and those who are deaf or hard of hearing may be removed by this technology. It can also help those who have speech or language impairments.

Recognition of sign languages is difficult because they are complex visual languages that depend on hand gestures, face emotions, and body movements. Additionally, each sign language has its own lexicon as well as grammar. As a result, developing sign language recognition systems necessitates a thorough understanding of computer vision principles and sign language linguistics. Despite this, scientists have made great strides in this field recently and are working to develop more accurate and dependable SLR systems for a variety of uses, including communication, entertainment, and education.

## 1.4    Bangla Sign Language(BdSL)

The deaf community in Bangladesh uses BdSL, also known as Bengali Sign Language, which is a visual language. This language, which has its own particular grammar, syntax, and vocabulary, is used to communicate through a mixture of gestures made with the hands and expressions on the face, and movements of body.

Around 2.6 million deaf people are currently residing in Bangladesh.[18] Despite the large quantity of individuals who are deaf or experiencing difficulty in hearing in Bangladesh, BdSL usage and acceptance have been constrained. Communication difficulties and social marginalization have been experienced by the deaf population as a result of inadequate resources and support. The Bangladesh Institute of Sign Language and Deaf Studies was founded as one of several projects to promote and boost the usage of BdSL.

It is crucial to spread awareness and encourage BdSL's development since they are necessary for the deaf people in Bangladesh's communication and social integration.

## 1.5    Bangla Sign Language(BdSL) Recognition

The field of research on Bangla Sign Language recognition is developing technology that can automatically interpret and translate Bangla Sign Language into text or speech. In Bangladesh, where there are a great number of deaf or hard of hearing persons, this technology has the potential to lessen communication barriers between those who utilize spoken language and those who are hard of hearing.

Bangla Sign Language is a sophisticated visual language that communicates meaning using hand gestures, facial expressions, and body language. Its grammar, syntax, and vocabulary are all distinct from those of other sign languages. Additionally, we can observe regional variations in Bangla Sign Language. For instance, a Sylhet resident using bangla sign language will use somewhat different signs than a Chittagong resident who uses BdSL. As a result, creating BdSL identification systems necessitates a thorough understanding of computer vision principles and sign language linguistics. Despite the difficulties, this discipline has made major strides recently as researchers attempt to develop more precise and reliable system of sign language recognition that can be used in communication, entertainment, and education.

The communication and social integration of Bangladesh's deaf people depend on the recognition and inclusion of BdSL. Therefore, it is crucial to raise awareness of and support for the creation of tools and technologies that can improve the understanding of Bangla Sign Language.

## 1.6   Problem Formulation

Given a video of sign language $V_i = \{f_1, f_2, \ldots, f_t\}$ with $t$ frames, our goal is to learn the conditional probabilities $p(S_i|V_i)$ of generating a translated sentence $S_i = \{w_1, w_2, \ldots, w_n\}$ with $n$ words.The training dataset consists of a set of tuples $\{(V_i, S_i) : i \in [1, P]\}$ where $P$ represents the total number of training videos.

There are quite a few challenges in solving SLT problem. Sign gestures can have different lengths because of it being done by different signers at different pace. Besides, as each video frame does not to map one to one with each of the tokens of translated sentences, the detailed meaning expressed through sign gestures might include subtle local aspects because of differences in grammatical rules and ordering in sign languages and spoken languages.

Because of transformer networks' effectiveness in modeling sequence-to-sequence tasks, recent methods on SLT employed this architecture[4]. However, it is important to recognize that the representation found in this architecture has its limitations. While transformer architecture excels at learning contextual relationships, this architecture struggles to capture the topological aspect of the joints of a human body which is equally important in sign language.

Sign language gestures that involve movements of human body joints are best represented through a spatio-temporal skeleton graph. We can take advantage of such a graph to learn a new set of feature representations. To bridge the limitation of transformer architecture, we leverage Spatio Temporal Graph Convolutional Network (STGCN) to extract the relationship between spatial and temporal features from the skeletal structure. Using only one type of feature representation from the architecute is limiting as together they can give both contextual and spatio-temporal information. To make the translation task more efficient, we incorporated both transformer architecture and STGCN architecture in our method to learn a better and more meaningful representation.

## 1.7   Motivation

The necessity for an efficient and dependable method of promoting communication between sign language users led to the development of a system for recognizing Bangla Sign Language from video data. The main form of communication for people with hearing loss is sign language, which can be difficult for people who don't know sign language to understand and translate. With the facilitation of more effective and efficient communication, the recognition and translation of Bangla Sign Language from video data has the potential to considerably enhance the lives of people with hearing impairments. Users of sign language will be able to participate more completely in society because of the proposed system, which will improve their ability to communicate clearly and effortlessly.

## 1.8   Objective

The objectives of this project are as follows:

- The primary objective of the proposed project is to make a system for Bangla Sign Language identification and translation. Sign language users will be able to communicate more effectively and efficiently because of the proposed system's ability to recognize and translate sign language, which will enhance their quality of life and allow them to take part more completely in society.

- The system will be created to reliably recognize and translate Bangla Sign Language utilizing a variety of variables, such as hand motion, body alignment, and facial expressions.

- The suggested system will have the ability to produce complete phrases in Bangla Sign Language, ensuring effective and smooth communication in this language.

- The suggested system will be able to manage phrases of various lengths, which is a crucial requirement for assuring successful and coherent communication in sign language.

## 1.9   Contributions

We summarize the main contributions of this thesis as follows:

- Architectural Fusion: We integrate both transformer and STGCN architecture to enhance our methods ability to extract meaningful representation by leveraging contextual and spatio-temporal information at both broader and fine-grained levels.

- Fusion Strategy: We explore different fusion strategies between these architecture to identify the most effective one.

- State-of-the-Art Performance: Our method provides a new state-of-the-art performance in gloss free sign language translation across RWTH-PHOENIX-2014T[3], CSL-Daily[12] and How2Sign[6] dataset to guide future research in the domain.

- Dataset Benchmarking: We conducted benchmarking on the BornilDB v1.0[14] dataset.

## 1.10   Outline

We discuss the theoretical underpinnings of our approach in chapter 2, with the transformer serving as the primary focal point. Discuss the works that have already been done on translating bangla and other languages in chapter 3. Chapter 4 discusses our reseach methodology. Here, we also talk about the datasets and evaluation matrices that we employed. The last topic we covered in this section was our future plan. Then, in chapter 5, we covered the findings we made using various datasets. Chapter 6 served as our final chapter.

# Chapter 2

# Preliminaries

## 2.1 Transformer

### 2.1.1 Encoder-Decoder Architecture

The Transformer model, similar to previous seq2seq models, has an architecture of encoder-decoder in which the encoder processes input iteratively and output is generated by decoder iteratively. Each encoder layer generates encodings with information about relevant parts of inputs, which are passed to the subsequent encoder layer as inputs. Similarly, the contextual information of the encodings is utilized by every decoder layer to produce the output sequence, utilizing attention mechanisms. The attention mechanism weighs the relevance of each input part to generate the output. An attention mechanism is implemented independently in the layer of each decoder that extracts information from the output of the decoder that came before and only then information is drawn via means of encodings. Both the layers of encoding and decoding include extra processing steps such as feed-forward neural networks, residual connections, and layer normalization to enhance the output.

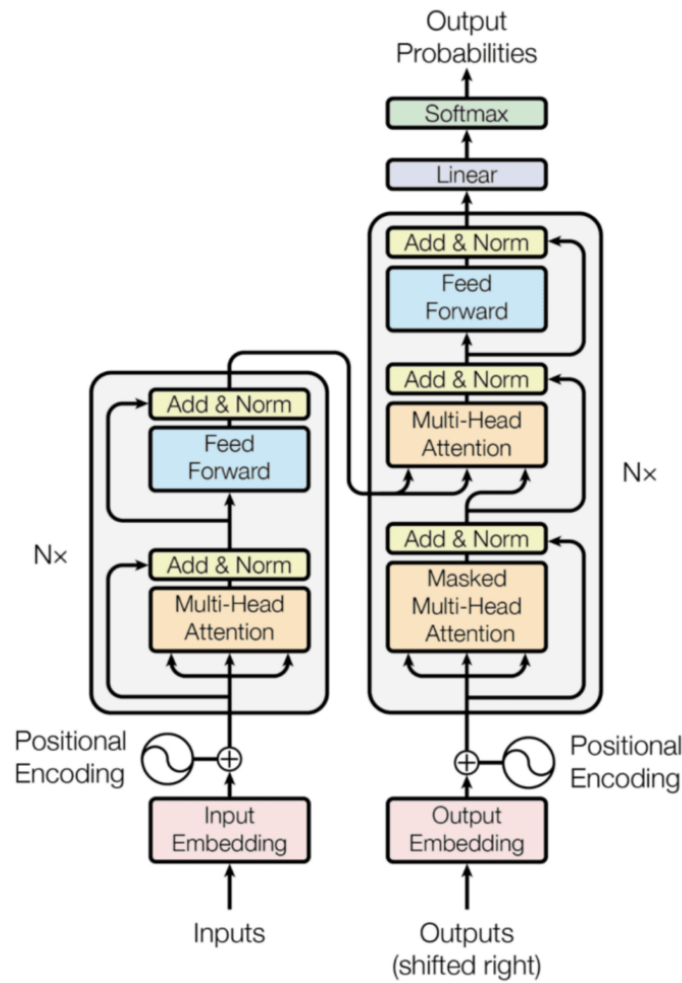

Figure 2.1: Basic Architecture of Trasformer[1]

### 2.1.2   Attention using the scaled dot-product method

The fundamental components of a transformer are units that use scaled dot-product attention method. Given as a sentence input through the transformer architecture, the weights of attention are computed concurrently all tokens in parallel. These units generate embedded contexts for each token, which comprise data about the token and a weighted sum of other significant tokens, in a way that each weight is determined by the attention weight.

The transformer model is equipped with three weight matrices for each attention unit, namely the weights of query $W_Q$, the weights of key $W_K$, and the weights of value $W_V$. The three weight matrices are multiplied by the input word embedding $x_i$ for each token $i$, generating a vector of query $q_i = x_i W_Q$, vector of key $k_i =$ $x_i W_K$, and vector of value $v_i = x_i W_V$. To compute attention weights, the vectors of query and key are used: the attention weight $a_{ij}$ from token $i$ to token $j$ is the dot product between $q_i$ and $k_j$. The square root of the dimension of the primary vectors is used to split these weights, $\sqrt{d_k}$, which improves gradient stability through the training, and then passed into a softmax function to normalize them. As $W_Q$ and $W_K$ are distinct matrices, attention can be asymmetrical. If token $i$ attends to token $j$ (i.e., $q_i \cdot k_j$ is high), this does not imply that token $j$ will attend to token $i$ . The weighted total of all tokens' value vectors, multiplied by $a_{ij}$, the attention from token $i$ to each token, is the result of the attention unit for token $i$.

By using softmax function, one huge matrix calculation can be used to represent the attention calculation for all tokens. This is advantageous for training since it allows for optimized matrix operations that can be computed rapidly. The matrices $Q$, $K$, and $V$ are defined as the matrices where the $i$th rows are vectors $q_i$, $k_i$, and $v_i$, respectively. Therefore, the attention can be represented as a matrix operation:

$$\text{Attention}(Q, K, V) = \text{softmax}\left(\frac{QK^T}{\sqrt{d_k}}\right) V$$

The horizontal axis is the axis over which the softmax function is applied.

### 2.1.2.1   Multi-Head Attention

In transformer models, a group of matrices $(W_Q, W_K, W_V)$ is called as attention head, and each layer contains multiple attention heads. The purpose of each of the attention head is to concentrate on these tokens are pertinent to a specific token, and with several attention heads, the task according to various explanation of "pertinence" is accomplished by the model. Moreover, the importance field that represents pertinence can become increasingly expanded across consecutive layers. Several attention heads in transformer models code meaningful pertinence relationships that are intuitive to normal humans. For instance, some of the attention heads may prioritize the next word, on the other hand others may primarily focus on the verbs and their direct objects. The calculations for each attention head can be executed simultaneously, enabling fast functioning. The output that is given by the attention layer are combined to form input for the layers of feed-forward neural network.

More specifically, if we index the multiple attention heads as $i$, then we obtain the following:

MultiheadedAttention$(Q, K, V) = $ Concat(Attention$(QW_i^Q, KW_i^K, VW_i^V))W^O$

In each attention head $i$, the matrices $W_i^Q, W_i^K, W_i^V$ are known as "projection matrices," while the final projection matrix $W^O$ is owned by the ultimate projection matrix that belongs to the whole multi-headed attention mechanism.

### 2.1.2.2    Masked Attention

Attention connections between certain word-pairs might be essential to eliminate. An instance of this could be the disconnection of the decoder from the token position $t + 1$. To achieve this, a mask matrix $M$ may be introduced prior to the softmax stage, whereby entries that require attention separation are assigned negative infinity, while other locations are assigned zero.

### 2.1.3    Encoder

The encoder in a transformer model comprises two primary elements: a self-attention mechanism and a feed-forward neural network. The self-attention mechanism utilizes preceding encoder's input encodings and their correlation with one another to create output encodings. Next, the feed-forward neural network handles every output encoding independently. The output encodings are subsequently transmitted on to the following encoder and the decoders.

During the first encoding stage, instead of encodings, inputs that consist of positional information and the input sequence's embeddings are provided. The positional information is essential to enable the transformer to utilize the sequence order since no other component of the transformer leverages this aspect.

### 2.1.3.1    Positional Encoding

A vector representation of specific size that sends a target sequence's tokens' relative positions is positional encoding. It provides information on how the order of the words in the input is established to the transformer model. The positional encoding can be described as a function $f : \mathbb{R} \to \mathbb{R}^d$, where $d$ is a positive even integer. According to the original paper's definition, the full positional encoding can be expressed by the equation below:

$$\text{PE}(pos, 2i) = \sin\left(\frac{pos}{10000^{2i/d_{\text{model}}}}\right) \tag{2.1}$$

$$\text{PE}(pos, 2i+1) = \cos\left(\frac{pos}{10000^{2i/d_{\text{model}}}}\right) \tag{2.2}$$

In the original paper, it was specified that the value of $d_{\text{model}}$ is equal to 512, which implies that the variable $i$ ranges between 0 and 255.

### 2.1.4   Decoder

The decoder component in a transformer model consists of three primary components, namely, a self-attention mechanism, a feed-forward neural network, and an attention mechanism over encodings. Decoder operates in such a way that is comparable to the encoding although it requires an additional attention mechanism that extracts relevant data derived from the encodings that were produced by the encoder. This mechanism is referred to as encoder-decoder attention. Just like the first encoder, the information of information and output sequence embeddings are received the first decoder as input, rather than encodings. In order to allow autoregressive text production and stop the reverse information flow, the output sequence is partially masked. Attention is not allowed to be placed on subsequent tokens for all attention heads. Finally, the probabilities of the vocabulary's output are generated by through the final linear transformation and softmax layer that are used on the last decoder output.

## 2.2   Graph

Graph is a set of vertices and edges represented by, $G = (V, E)$ where $V$ is the set of vertices or nodes and $E$ is the set of edges. Any node in the graph, G, defined as $v_i \in V$ and edges as $e_{ij} = (v_i, v_j) \in E$ denote that there exists an edge between node $v_i$ to $v_j$. In case of undirected graphs those edges exist in both direaction. Let the neighborhood of a node $v$ define as $N(v) = \{v \in V | (v, u) \in E\}$. The adjacency matrix is a $n \times n$ matrix where $n$ denote number of nodes. The first hop adjacency matrix is $A_{ij} = 1$ if $e_{ij} \in E$ and $A_{ij} = 0$ if $e_{ij} \notin E$. Let $d(v_i, v_j)$ is an edge counting function which count minimum number of edges traversed from node $v_i$ to node $v_j$. The adjacency matrix can also be defined with the help of edge counting function $d(v_i, v_j)$. The first hop adjacency matrix defined as $A_{ij} = 1$ if $d(v_i, v_j) = 1$ and $A_{ij} = 0$ if $d(v_i, v_j) > 1$. The k-hop adjacency matrix is $A_{ij} = 1$ if $d(v_i, v_j) \leq k$ and $A_{ij} = 0$ if $d(v_i, v_j) > k$. The degree of a node is the number of edges connected with that node. The degree matrix is a $n \times n$ matrix with $D_{ij} = \sum_{k=0}^{n-1} A_{ik}$ if $i = j$ and $D_{ij} = 0$ if $i \neq j$. The graph Laplacian is, $L = D - A$. The normalize graph Laplacian is $\mathcal{L} = D^{-1}L = D^{-\frac{1}{2}}D^{-\frac{1}{2}}L = I - D^{-\frac{1}{2}}AD^{-\frac{1}{2}}$.

## 2.3    Spatial Graph Convolution

### 2.3.1    Neural Message Passing

The essential notion of the graph neural network (GNN) model is a versatile convolutional framework suitable for non-Euclidean settings[19], analogous to traditional graph isomorphism test[20]. Regardless of the motive, a GNN's distinguishing feature is that it employs message passing between neighbourhood nodes and update this information using neural network similar to neural message passing[21].

### 2.3.2    The Framework for Passing Messages

In each iteration of message-passing within a GNN, a hidden representation denoted as $h_u^{(k)}$ for each node $u$ in the set $\mathcal{V}$ is adjusted based on information gathered from the neighborhood of node $u$, represented as $\mathcal{N}(u)$ (as illustrated in Figure 2.2). This update during message-passing can be described as follows:

$$h_u^{(k+1)} = \text{UPDATE}^{(k)}\left(h_u^{(k)}, \text{AGGREGATE}^{(k)}(\{h_v^{(k)}, \forall v \in \mathcal{N}(u)\})\right) \quad (2.3)$$

$$= \text{UPDATE}^{(k)}\left(h_u^{(k)}, m_{\mathcal{N}(u)}^{(k)}\right) \quad (2.4)$$

Here, the terms UPDATE and AGGREGATE refer to flexible differentiable functions, typically implemented as neural networks. The variable $m_{\mathcal{N}(u)}$ represents the message collected from the graph neighborhood $\mathcal{N}(u)$ of node $u$. We utilize superscripts to distinguish between the embeddings and functions employed at various stages of message passing iterations.

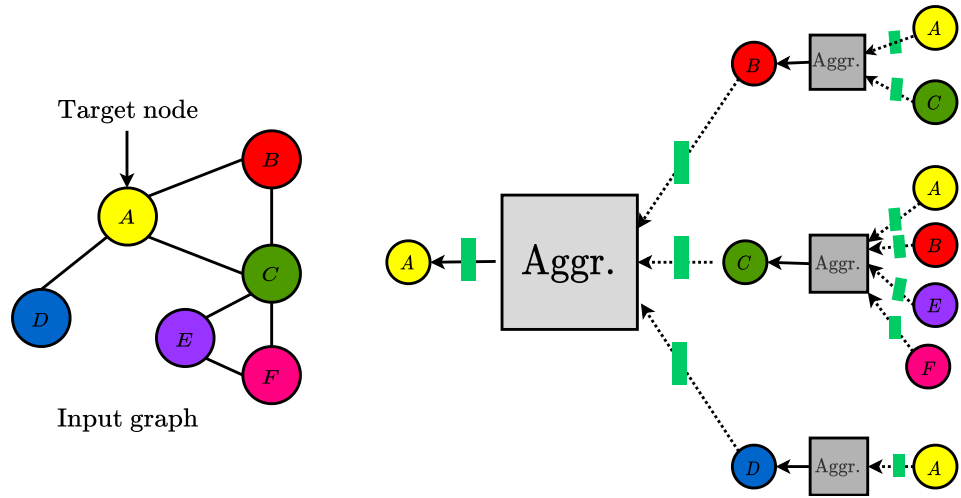

Figure 2.2: A visual representation that demonstrates the process of a single node gathering information from its nearby neighbors. The model collects information from local graph neighbors of node $A$ (specifically, neighbors $B$, $C$, and $D$), and the content of these messages is built upon information that those neighbors have gathered from their own respective neighborhoods, forming a cascade of information aggregation. A two-layer message-passing mechanism is represented in this diagram. By expanding the neighborhood around the target node, the GNN computation graph generates a tree shape. The green bar indicates feature transformation and Aggr. means feature aggregation.[2]

### 2.3.3 Basic Graph Neural Network

The GNN framework has been outlined in a general manner as a series of message-passing steps that make use of the UPDATE and AGGREGATE functions. We need to provide specific and clear definitions of UPDATE and AGGREGATE functions in order to implement this idea of message passing. We will start with the most simplified version of the GNN framework, which inspired by the original GNN models [22], [23]. The fundamental GNN message passing is described as:

$$h_u^k = \sigma\left(\boldsymbol{W}_{\text{self}}^{(k)} h_u^{(k-1)} + \boldsymbol{W}_{\text{neigh}}^{(k)} \sum_{v \in \mathcal{N}(u)} h_v^{(k-1)} + b^{(k)}\right) \quad (2.5)$$

Here, we have matrices $W^{(k)}$self and $W^{(k)}$neigh as trainable parameters, both with dimensions in $\mathbb{R}^{d^{(k)} \times d^{(k)}}$. The symbol $\sigma$ represents an elementwise non-linear function, such as tanh or ReLU. It's worth noting that the bias term $b^{(k)}$ in $\mathbb{R}^{d^{(k)}}$ is often left out for simplicity in notation, but including this bias term can be essential for achieving improved performance.

The basic GNN framework's message passing is similar to that of a multi-layer perceptron (MLP). We aggregate the messages received from its neighbors; next, using a linear combination, we merge the information from the node's surroundings with the node's prior embedding; and finally, we apply elementwise non-linearity.

### 2.3.4   Passing Messages with Self-Connections

Introducing self-loops by adding it into the input graph is a common practice to streamline the GNN message passing process. In this methodology, the transmission of messages is straightforwardly defined as:

$$h_u^{(k+1)} = \text{AGGREGATE}\bigg(\{h_v^{(k)}, \forall v \in \mathcal{N}(u)U\{u\}\})\bigg) \tag{2.6}$$

where now the aggregation of this self-loop is overpowered over the set $\mathcal{N}(u)U\{u\}$ i.e., both the neighbors of the node and the node itself. The advantage of this strategy is that we don't need to specify an explicit update function because the update is defined implicitly by the aggregation method. Simplifying the message passing in this manner can frequently reduce the risk of overfitting but it drastically restricts the expressivity of the GNN because information from rhe neighbors of the node are not allowed to be distinguished the information from itself.

For the fundamental GNN, when self-loops are introduced, it's akin to having shared parameters between the Wself and Wneigh matrices, as indicated in Equation 2.5. This results in the subsequent update at the graph level:

$$\boldsymbol{H}^t = \sigma\bigg((\boldsymbol{A} + \boldsymbol{I})\boldsymbol{H}^{(t-1)}\boldsymbol{W}^{(t)}\bigg) \tag{2.7}$$

### 2.3.5   Invariance and Equivariance with Respect to Permutations

A practical method for constructing a DNN for graphs is to directly input the adjacency matrix into the network. For instance, when aiming to create an embedding for the entire network, one approach could involve flattening the adjacency matrix and then passing the result through a multi-layer perceptron (MLP).

$$z_g = \text{MLP}(\boldsymbol{A}[1] \oplus \boldsymbol{A}[2] \oplus ... \oplus \boldsymbol{A}[\mathcal{V}])$$

Here, $\boldsymbol{A}[i] \in \mathbb{R}^{|\mathcal{V}|}$ represents a row in the adjacency matrix, and we use the symbol $\oplus$ to signify vector concatenation.

The problem with this method lies in its reliance on the specific node order we use in the adjacency matrix. In simpler terms, this model lacks the property of permutation invariance, which is crucial when designing neural networks for graph-related tasks. Ideally, any function, denoted as $f$, that takes an adjacency matrix $A$ as input should possess the following characteristics:

$$f(PAP^T) = f(A) \qquad\qquad \text{(Permutation invariance)}$$
$$f(PAP^T) = Pf(A) \qquad\qquad \text{(Permutation equivariance)}$$

Here, $P$ is a permutation matrix. A function is considered permutation invariant if it remains unchanged irrespective of the arbitrary arrangement of columns and rows in the adjacency matrix. On the other hand, permutation equivariance means that the output of the function $f$ is consistently rearranged when we permute the adjacency matrix. Ensuring either invariance or equivariance is a significant challenge when dealing with graph-based learning.

## 2.4   Spectral Graph Convolution

### 2.4.1   Laplacian smoothing

Laplacian Smoothing is a technique used to apply regularization on a graph's relational structure, primarily applied in the context of node classification in a semi-supervised way. In the context of Laplacian regularization, we define a function that depends on the graph of interest and the space to be smoothed. This function can be expressed as:

$$LS(P, f(X)) = \sum_{(i,j) \in E} ||f(X_i) - f(X_j)||^2 = f^T \Delta(P) f \qquad (2.8)$$

Here, $\Delta$ represents the graph Laplace operator, and $LS(P, f(X))$ is Laplacian Smoothing applied to the target space, denoted as $f(X)$, with $L = \Delta(P)$. The proximity matrix $P$ is a user-defined matrix of dimensions $n \times n$. Laplacian smoothing can be applied to various spaces, such as the label space (where $f(X) = L$), data projections $X$, or a latent cluster space (where $f(X) = C$), as proposed in our work.

The loss function is formulated as:

$$Loss = Loss_{supervised} + \lambda LS(P, f(X)) \qquad (2.9)$$

In this formulation, the regularization term is incorporated into the learning objective and optimized alongside the supervised loss. The parameter $\lambda$ in the equation represents a trade-off parameter, which balances the importance of neighborhood similarity with node features. It is typically determined through cross-validation. Notably, Laplacian SVM [24] is a well-known model that combines a Laplacian regularizer with the Hinge loss. Additionally, one can consider Label Propagation as a form of Laplacian smoothing applied to the label space, utilizing a k-nearest neighbor classifier.

### 2.4.2   Applying spectral kernels to convolutions

The normalized Laplacian matrix can be decomposed using eigenvalues as $\mathcal{L} = U \Lambda U^T$. Here, $U$ is a matrix of eigenvectors ordered by their corresponding eigenvalues, and $\Lambda$ is a diagonal matrix of these eigenvalues, where $\Lambda_{ii} = \lambda_i$, and $\Lambda_{ij} = 0$ when $i$ and $j$ are not equal. Since the Laplacian matrix, L, is symmetric, its eigenvector matrix forms an orthonormal basis that satisfies the property $UU^T = I$. The graph Fourier transform of a signal $x$ is defined as $f(\Lambda) = U^T x$, and the inverse graph Fourier transform is defined as $f(x) = U f(\Lambda)$.

Now the graph convolution of the input signal $x$ with a filter $g \in R^n$ is defined as:

$$x * g = U(U^T x \odot U^T g) \tag{2.10}$$

where $\odot$ is the elementwise product. Now we will denote the filter as $g_\theta = diag(U^T g)$, then the spectral graph convolution is:

$$x * g_\theta = U g_\theta U^T x \tag{2.11}$$

All Spectral-based Convolutional Graph Neural Networks (ConvGNNs) adhere to the previously mentioned definition. However, the main distinction lies in the selection of the filter, denoted as $g_\theta$.

In the context of the Spectral CNN, the filter denoted as $g_\theta$ is populated with a set of adjustable parameters, specifically represented as $\theta_{i,j}^k$. The graph convolutional layer in the Spectral CNN is established as follows:

$$H_{:,j}^k = \alpha(\sum_{i=1}^{f_{k-1}} U \Theta_{i,j}^k U^T H_{:,i}^{k-1})$$

where $k$ is the layer index, $H^{k-1} \in R^{n \times f_{k-1}}$ is the input graph signal, $H^0 = X$, $f_{k-1}$ is the number of output channels, $\Theta_{i,j}^k$ is a diagonal matrix filled with learnable parameters.

Spectral CNNs have certain limitations, with one notable drawback being the high computational complexity of eigenvalue decomposition, which requires a time complexity of $O(n^3)$. In subsequent works like ChebNet [25] and GCN [26], efforts have been made to reduce this computational burden to $O(m)$ by introducing various approximations and simplifications.

For instance, Chebyshev Spectral CNN (ChebNet) utilizes Chebyshev polynomials to estimate the filter $g$, where $g = \sum_{i=0}^{k} \theta_i T_i$, and $\tilde{\ }$ is constrained within the range [-1,1]. It's important to recall the recursive definition of Chebyshev polynomials: $T_i(x) = 2x T_{i-1}(x) - T_{i-2}(x)$, with initial values of $T_0(x) = 1$ and $T_1(x) = x$. As a consequence, the convolution of a graph signal x with the specified filter $g$ can be represented as:

$$x * g_\theta = U(\sum_{i=0}^{k} \theta_i T_i(\tilde{\Lambda}))U^T x = \sum_{i=0}^{k} \theta_i T_i(\tilde{L})x$$

The Graph Convolutional Network (GCN) [26] employs a simplified version of the ChebNet, specifically the first-order approximation. For the sake of this discussion, let's assume that the values of $k$ and $\lambda_{max}$ are set to 1 and 2 respectively.

$$x * g_\theta = \theta_0 x - \theta_1 D^{-\frac{1}{2}} A D^{-\frac{1}{2}} x$$

In order to decrease the parameter count and mitigate the risk of overfitting, GCN takes an additional assumption that $\theta$ can be simplified as $\theta = \theta_0 = -\theta_1$. This assumption leads to the subsequent definition of graph convolution:

$$H = x * g_\theta = \sigma(\tilde{A} X \Theta) \tag{2.12}$$

where $\tilde{A} = D^{-\frac{1}{2}}(A + I_n)D^{-\frac{1}{2}}$ and $\sigma$ is the nonlinear activation function. Due to the self-influence identity matrix, $I_n$, is added with the adjacency matrix, $A$.

We can view GCN as transforming feature matrix, $X$, with a learnable linear transformation matrix, $\Theta$, then aggregative the neighbourhood information using the adjacency matrix, $A$. We can further improve the model by parameterizing the aggregation step and then learning those parameters. Graph Attention Networks [27] introduce such an attention mechanism where we learn how to set priority on each node. The graph convolutional operation as per the Graph Attention Network (GAT) is specified as follows:

$$h_v^k = \sigma(\sum_{v \in N(v)Uv} \alpha_{uv}^k W^k h_v^{k-1}) \tag{2.13}$$

where $h_v^0 = x_v$. The attention weight $\alpha_{uv}^k$ measure the connective strength between the node $v$ and its neighbour $u$.

$$\alpha_{uv}^k = \sigma(\mu(a^T[W^k h_v^{k-1} || W^k h_u^{k-1}])) \tag{2.14}$$

where $a$ is a vector of learnable parameters, $\sigma(\cdot)$, $\mu(\cdot)$ represent sigmoid and leakyrelu respectively. The softmax function guarantees that the attention weights collectively add up to one when considering all neighboring nodes of the node $v$.

# Chapter 3

# Related Work

Initial research on Sign Language Recognition (SLR) primarily focused on isolated SLR [28, 29, 30, 31, 32, 33, 34], which identifies individual signs but lacks the natural flow of sign language communication. Limitations of isolated SLR spurred the development of continuous SLR [35, 36, 37, 38, 39, 40, 41], that aims to continuously recognize sign gestures, aligning with sign language grammar and structure, catering to the preferences of sign language users.

In pursuit of rfficient communication between individuals proficient in sign language and those who are unfamiliar with sign language, extensive research has been done on continuous SLT. Initially, SLT employed Recurrent Neural Networks (RNNs) [42] within the encoder-decoder framework, utilizing either Gated Recurrent Units (GRUs) or Long Short-Term Memory (LSTM) [43, 44, 45, 46]. However, addressing the limitations of RNNs in handling long-term dependencies has led to the adoption of more effective attention-based methods. The transformer network[1], known for its success in various domains[47, 48, 49, 50, 51, 52], derives its efficacy from its self-attention mechanism, a feature that has found favor in recent SLT research, leading to the widespread adoption of the transformer architecture. But transformers can only capture and learn the contextual information and patterns. It cannot take advantage of the inherent Graph structure made up of joints in a human body. Spatial-temporal graph convolutional networks (STGCN) potential to learn spatio-temporal dynamics have been demostrated in Human Activity Recognition[53]. Consequently, this architecture has been employed to capture spatial and temporal relationships that is essential for SLR[54, 55]. However, fusion of contextual and spatio-temporal relationships using both transformer and STGCN architecture have not been explored, which overlooks important aspects of sign language. In our method, we leveraged both these architectures to get a better and meaningful representation of the sign gestures.

Sign language translation based on gloss supervision categorizes current SLT methods into three groups: two-stage gloss-supervised methods, end-to-end gloss-supervised methods, and end-to-end gloss-free methods. Gloss, a textual representation of sign gestures in spoken or written language, serves as an intermediary in the first two approaches[3, 4, 12, 56]. However, acquiring gloss annotations can be costly as it requires the expertise of sign language professionals. In contrast, gloss-free methods do not rely on intermediary gloss annotations, directly translating sign language videos into spoken language texts[46, 57, 58, 59, 60]. As seen from the work of [7] gloss annotation helps with alignment information which can supervise the attention calculation to focus on the correct positions of a video. This supervision is lost without the help of gloss when traditional attention mechanisms are used ,reducing performance of the models without the supervision of gloss drastically. So there is a trade-off between the cost to acquire annotation and translation performance. Our method belongs to the gloss-free category as we utilize a direct translation from sign language video to spoken language text without any involvement of gloss annotation.

Various SLT approaches have employed different tokenization methods for sign language videos. Some utilized 2D CNN features extracted from video frames at the gloss-level [3, 4]. Inflated 3D convnets (I3D), initially designed for action recognition [61], have been further trained using sign language data [5, 46, 59, 62, 63, 64]. S3D [65] features have been employed after pretraining with the WLASL dataset and kinetics [66]. Additionally, some approaches have used pose estimators[67, 68] to represent video sequences, as they provide information on the position and movement of body parts[44, 45, 58, 69]. Lastly, some methods combined both video and keypoint feature to capture more meaningful representation[56, 70]. Our method aligns with this last mentioned method as we use both rgb video encoding and keypoint encoding.

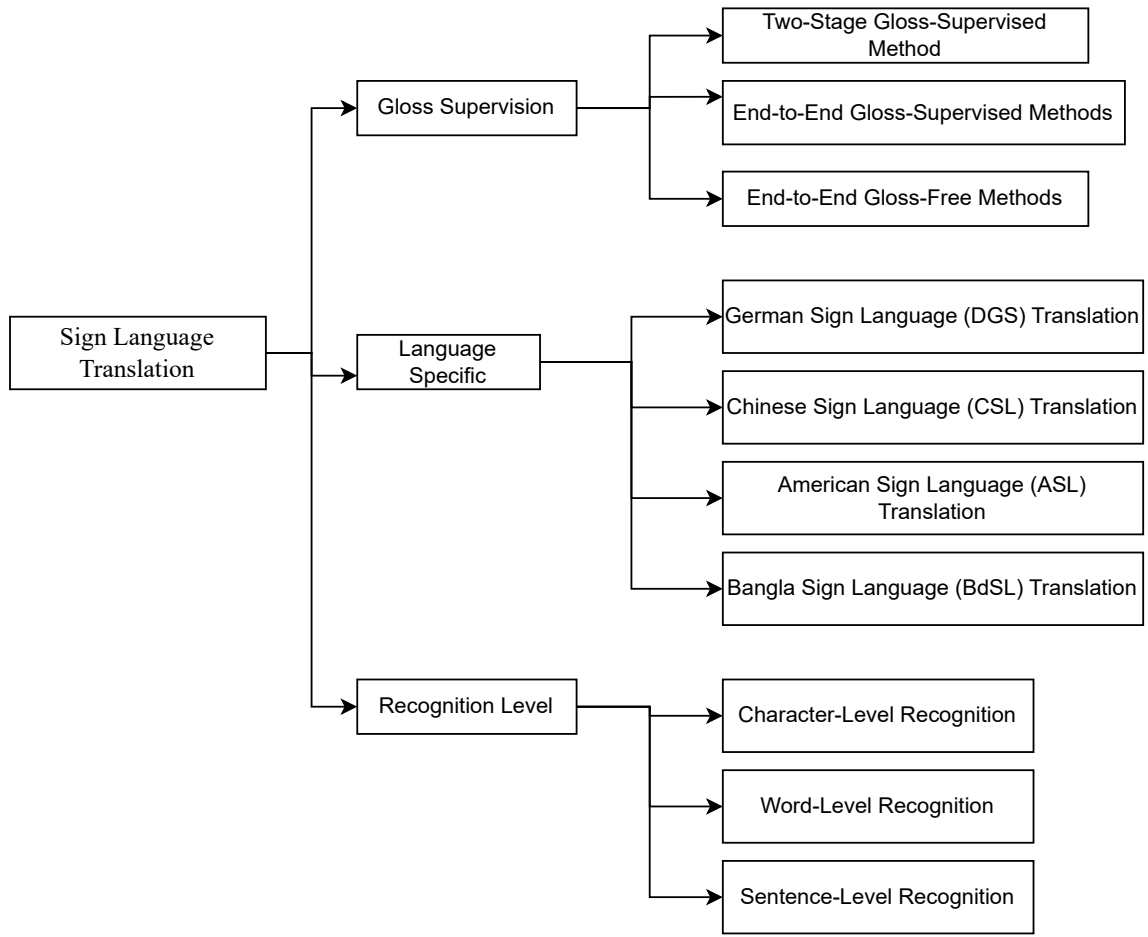

Figure 3.1: Variations of Sign Language Translation [drawn in draw.io]

## 3.1   Two-Stage Gloss-Supervised Methods

Camgoz et al. [3] proposes a novel approach to solve the problem of SLR and
SLT using deep learning techniques. Unlike previous SLR research that focuses
on basic gesture recognition, the proposed method treats the task just like neural
machine translation (NMT) problem, incorporating linguistic features of sign lan-
guage. The approach involves learning the spatio-temporal representation of signs,
their relation, and how they translate to spoken or written form using sequence-to-
sequence deep learning techniques. The paper introduces a novel vision methods
that replicate the comparable to the tokenization and embedding processes uti-
lized in conventional NMT and presents RWTH-PHOENIX-Weather 2014T, the
first continuous SLT dataset. The contributions included the first exploration of
the video-to-text SLT problem, the introduction of the first publicly available con-
tinuous SLT dataset, and a benchmark results on the new corpus. The outcomes
consist of diverse techniques for tokenization and attention, along with suggestions
for parameter settings.

The paper presents a novel end-to-end solution for SLT, which recognizes sign
language as a unique language and underscores the significance of SLT. Also the
authors collected and publicly shared the first continuous sign language translation
dataset.

The analysis reveals that the predominant type of error is the incorrect inter-
pretation of numerical values, dates, and locations. Besides, a restriction of the
Sign-to-Text (S2T) network is its attention that is primarily focused at the begin-
ning of the video, and It solely moves to the end position only when processing
the final words. Despite the intermediate tokenization in the Sign-to-Gloss-to-Text
(S2G2T) network reducing the asynchronicity between sign channels, it still shows
many-to-one mappings. Additionally, both models have a constraint in capturing
the contextual meaning of certain spoken words that are not explicitly signed and
require interpretation via context.

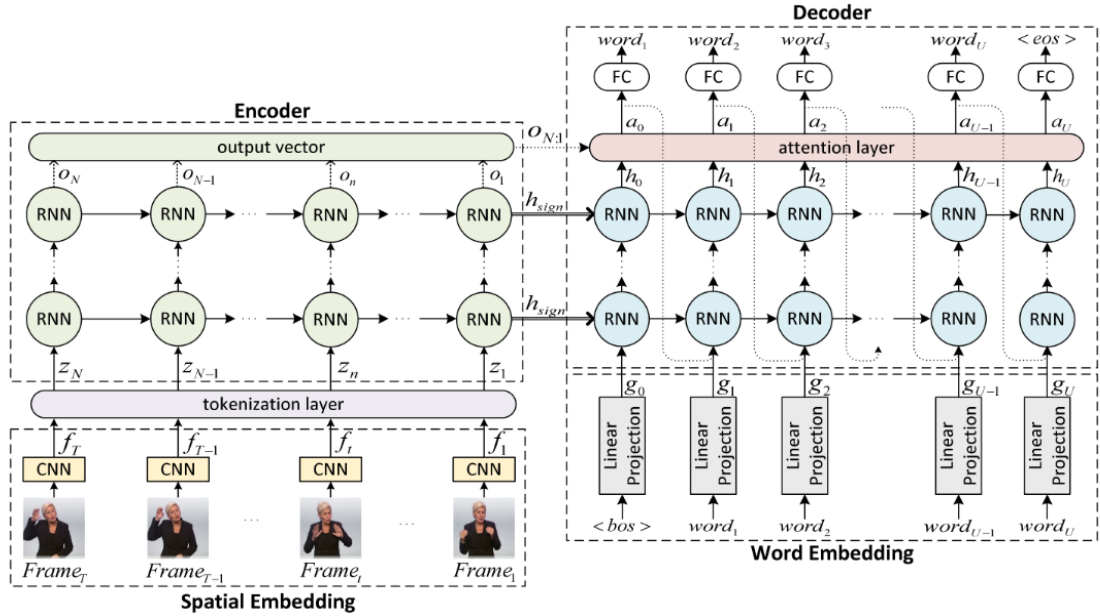

Figure 3.2: A summary of a method for producing spoken language translations of videos containing sign language[3]

## 3.2 End-to-End Gloss-Supervised Methods

A novel approach was presented by Camgoz et al.[4] for accomplishing both sign language recognition and translation simultaneously using two transformer models: the Sign Language Recognition Transformer (SLRT) and the Sign Language Translation Transformer (SLTT). The SLRT is an encoder transformer model that is trained with a CTC loss and is responsible for predicting sign gloss sequences. To achieve this, this method takes spatial embeddings derived from sign language videos and uses them to train the model to learn spatio-temporal representations. These representations are then utilized by the SLTT, which is an autoregressive transformer decoder model that predicts one word at a time and generates the corresponding spoken language sentence. They introduced a new way of approaching CSLR and SLT that involves multiple tasks,transformers were effectively employed for CSLR and SLT, and a range of new baseline results.

The contributions of this paper include that the translations exhibit high quality and effectively convey the same meaning, despite differences in terminology. Also, the formed phrases demonstrate minimal deviation from standard grammar.

The most difficult items to translate are those that are named and have insufficient contexts in the training samples, such as places. Particular numbers also pose a challenge, as No grammatical context is present to differentiate them from each other.

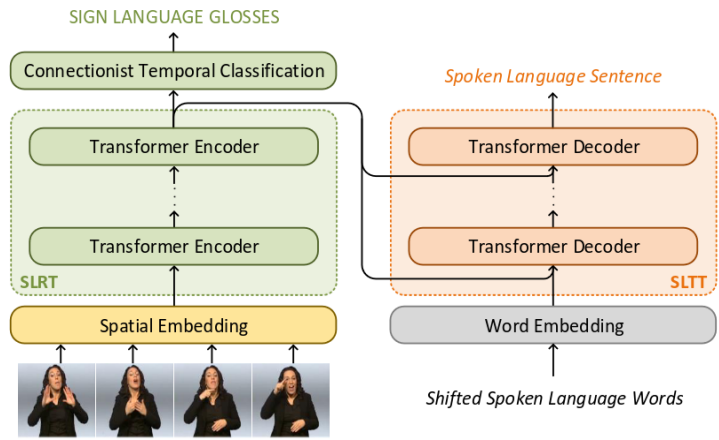

Figure 3.3: A summary of the method for performing Sign Language Recognition and Translation using transformers in an end-to-end manner.[4]

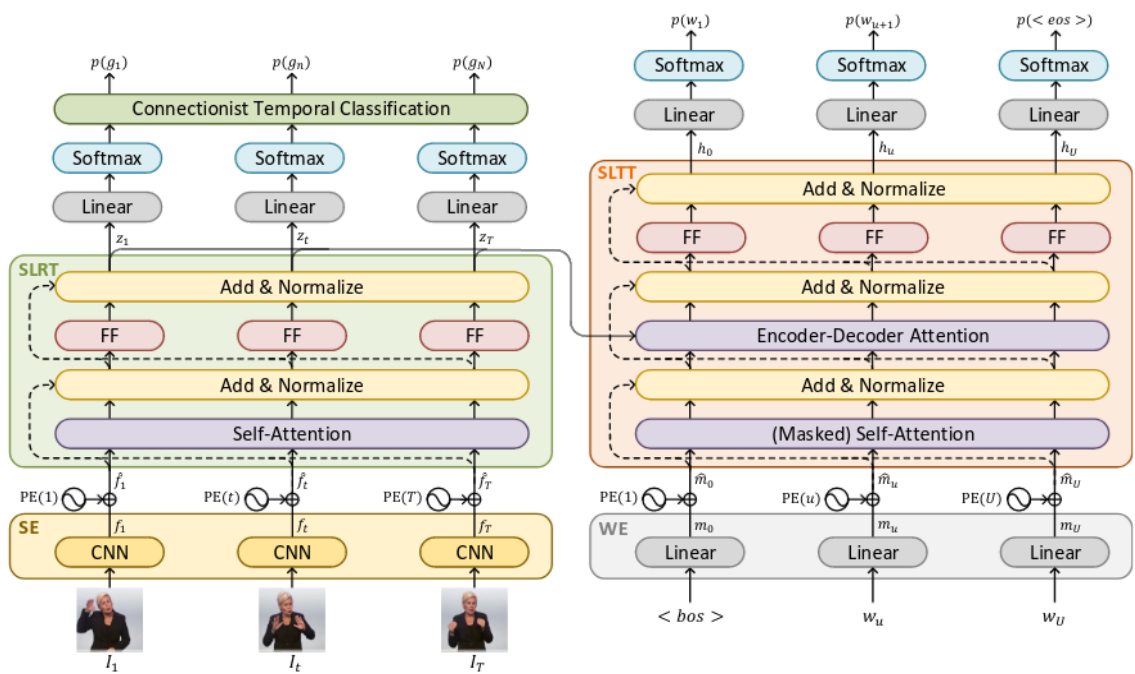

Figure 3.4: An elaborate explanation of a Sign Language Transformer with a single layer, which includes Spatial Embedding (SE), Word Embedding (WE), Positional Encoding (PE), and Feed Forward (FF) components.[4]

## 3.3   End-to-End Gloss-Free Methods

Li et al[5]. proposed a novel framework for doing sign language translation using a hierarchical feature learning method. A temporal semantic pyramid is used in the process to extract local and global temporal context data from sign language films. More expressive sign video features are produced by TSPNet's inter- and intra-scale attention mechanisms, which are used to learn features from noisy video clips. Tests show that TSPNet performs significantly better than earlier bootstrapping models, demonstrating the viability of the suggested strategy.The TSPNet model was assessed using the RWTH-PHOENIX-Weather 2014T. (RPWT) dataset.

The proposed framework presents the potential to learn sign language translation models directly from natural language sources, such as subtitled television footage, without the need for costly annotations. This significantly reduces the requirements for expensive annotations. The architecture is not limited to a specific backbone and may be applicable to other tasks, including weakly supervised action localization.

In their approach, the authors do not explicitly model the extent of signs, such as differentiating between "shower" and "rain storm," which is often reflected in facial expressions. The translation of low frequency words, such as city names, presents a significant challenge.

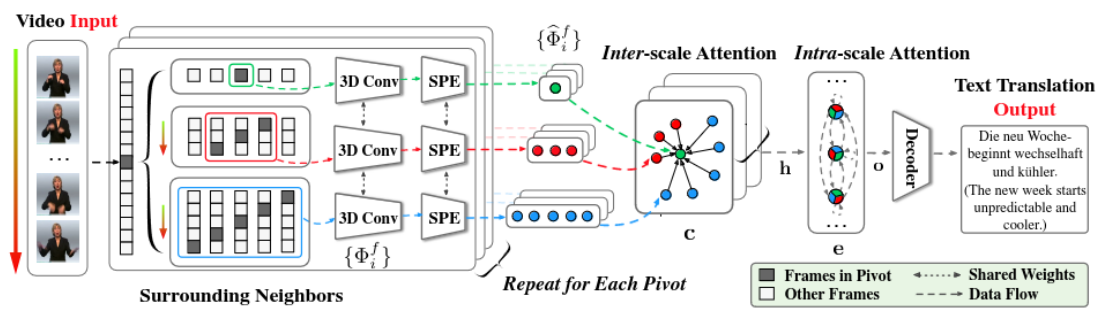

Figure 3.5: Overview of the proposed TSPNet workflow, which directly produces spoken language translations from sign language videos[5]

The task of translation of sign language is intricate and involves the conversion of continuous sign language videos into sentences in spoken language. Despite recent advancements in SLT, the problem remains unsolved, impeding effective communication between signers and non-signers. Limited availability of public sign language datasets, especially those with parallel video and textual translations for benchmarking, hinders progress in the field. The study[6] presents baseline SLT models trained on this dataset, attaining a BLEU-4 value of 8.03. The paper also highlights the limitations of solely relying on BLEU metrics for model selection and proposes the use of reduced BLEU (rBLEU) as an alternative evaluation metric.

The authors propose the use of reduced BLUE (rBLUE) as an alternative metric of BLUE for better characterizing the performance of sign language translation systems and selecting more appropriate checkpoints during training. The paper presents the very first sign language translation baselines on the dataset of How2Sign. By developing along with evaluating these baselines, the authors establish initial performance benchmarks for the task of translating ASL to written English using this dataset. The reported BLUE score of 8.03 illustrates the efficacy of the model.

The decoder component in the system under investigation appears to disregard the conditioning aspect given by the encoder, operating primarily as a language model rather than effectively utilizing the encoded information. Insufficient attention has been given to the exploration and investigation of optimal techniques for extracting and utilizing visual features specifically made for the task of translation of sign language.

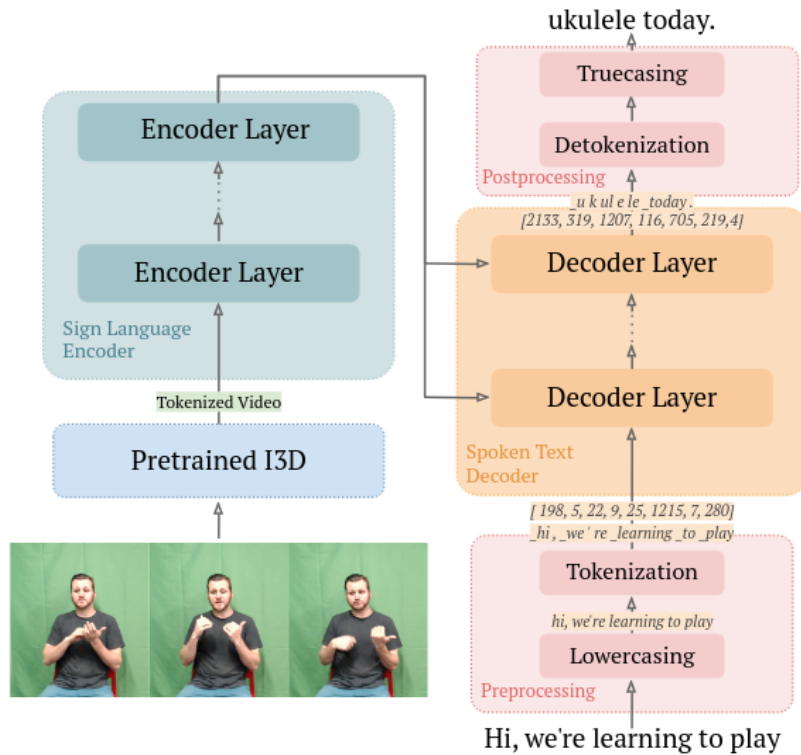

Figure 3.6: A Transformer is utilized to handle the sequence of input video and generate the corresponding sequence of output text.[6]

In another work, Yin et al[7] presents Gloss Attention Sign Language Translation (GASLT), a novel methodology designed to address the challenges associated with the utilization of gloss annotations in SLT. The study conducts an in-depth analysis of existing models and highlights two primary advantages conferred by gloss annotations: facilitating the acquisition of semantic boundaries within continuous sign language videos and promoting a comprehensive understanding of the global context. To mitigate the complexities surrounding the acquisition of gloss annotations, the GASLT model incorporates a gloss attention mechanism. This mechanism effectively directs the model's attention towards video segments exhibiting similar semantic properties, thereby simulating the role of gloss annotations. Empirical evaluations performed on large-scale sign language datasets substantiate the significant performance enhancements achieved by the GASLT model in comparison to established methods. Consequently, GASLT presents a highly promising solution by alleviating the dependency on gloss annotations within SLT systems.

This study conducts a comprehensive investigation and examination of the utilization of gloss annotations in the context of sign language translation. The research delves into an in-depth exploration and analysis of the application of gloss annotations, focusing on their role and impact within the domain of sign language translation. This study introduces a novel approach that eliminates the reliance on gloss annotations to some extent in sign language translation. The proposed method incorporates a new attention architecture and leverages knowledge transfer techniques to partially replace the need for gloss annotations.

Exhibits a relative decrease in performance compared to the state-of-the-art architectures and methodologies in gloss-supervised SLT. The model falls short in achieving comparable levels of accuracy and effectiveness as those achieved by the leading approaches currently employed in the field.

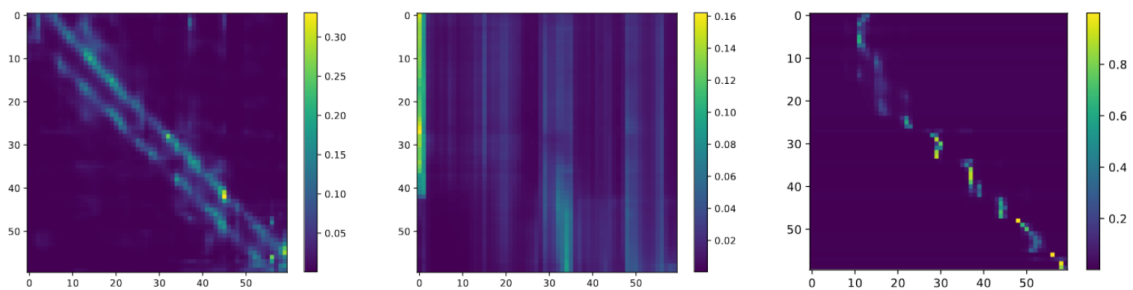

Figure 3.7: Visualisation of the attention map in three different SLT models' shallow encoder layers. As seen in (a), one of gloss's key functions is to give the model alignment information so that it can concentrate on substantially more crucial local areas. The conventional method for calculating attention faces difficulty in accurately converging to the correct position when the supervision signal of the gloss is lost, as illustrated in (b) By incorporating an inductive bias, which might be somewhat substitute for the function of gloss, our suggested approach enables (c) To maintain adaptable focus on critical regions, akin to how it's done in (a).[7]

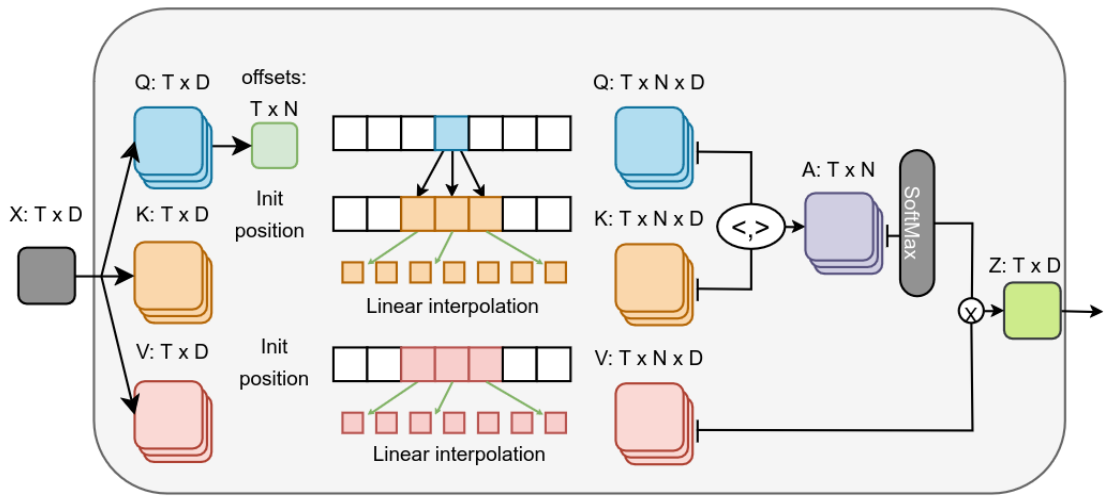

Figure 3.8: The flowchart of gloss attention involves an initial emphasis on a set of neighboring queries surrounding a specific query. Subsequently, the model calculates adjustments determined by the query to dynamically adjust the focus position. This dynamic adjustment ensures that the model maintains attention on the appropriate position, similar to how it operates with gloss supervision. Finally, a linear interpolation is employed to obtain the ultimate attention key and value. By performing softmax operations over the last dimension, the model achieves the desired attention distribution.[7]

Sign language recognition and translation also can be approached from three distinct but interconnected perspectives: character-level recognition, word-level recognition and sentence-level translation. In the former, the focus lies on the intricate and expressive characters that form the backbone of sign languages. This approach delves deep into the nuances of individual signs, their meanings, and the gestures and expressions that make each character unique. On the other hand, sentence-level translation broadens the scope, aiming to mitigate the rift between sign languages and spoken or written languages, allowing for comprehensive communication.

## 3.4   Character-Level Recognition

Numerous studies on the recognition of numerals and sign alphabets from images have been carried out in the context of the Bangla Sign language[8, 9, 71, 72, 73]. Urmee et al [8] created BdSLInfinite datatset, consisting of 2000 images and 37 distinct signs. They used transfer learning to solve this classification task based on a CNN architecture,Xception net.They achieved an accuracy of impressive 98.93% accuracy with an average response time of 48.53 ms. Notably, this method surpasses existing BdSL recognition approaches in terms of both accuracy and speed, making it a significant contribution to the field.

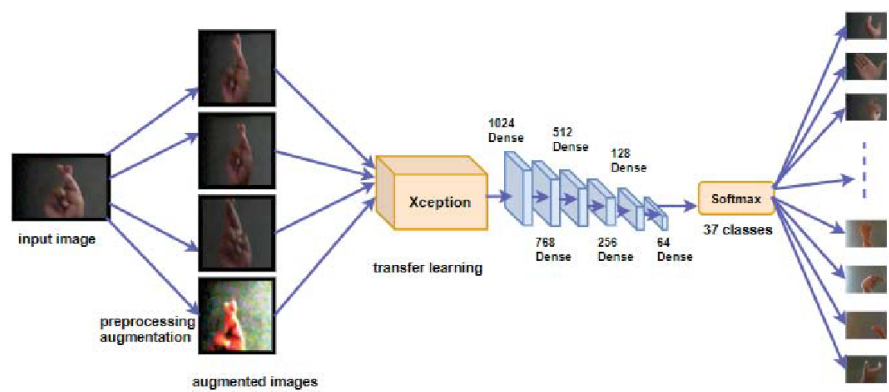

Figure 3.9:   Augmented images undergo transfer learning in the Xception model, which then makes predictions across 37 different classes[8]

Hoque et al [9] addressed the challenge of real-time detection of BdSL signs using a Convolutional Neural Network (CNN)-based object detection technique, Faster R-CNN. The authors introduced the BdSLImset dataset, featuring images of BdSL signs with diverse backgrounds and lighting conditions. Their method, which leverages Faster R-CNN, successfully identifies and recognizes BdSL signs in real-time without relying on external devices, outperforming previous techniques. Key contributions include the creation of the BdSLImset dataset and the development of a real-time BdSL sign detection system using Faster R-CNN.

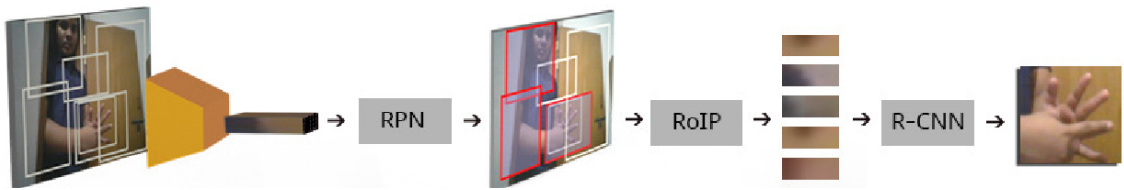

Figure 3.10: The network's intricate structure starts with the input image entering the CNN framework, generating a feature map. Subsequently, the Region Proposal Network (RPN) identifies anchors with a greater likelihood of being valid, and the final stage involves RoI pooling for classification, ultimately completing the detection process.[9]

## 3.5 Word-Level Recognition

A team of undergraduate thesis students at BRAC University conducted research on the text generation of BdSL. Their study[10] utilized a dataset consisting of 34 words in Bangla sign language created by them. The methodology involved preprocessing the video input to obtain training data, followed by the use of classification models (specifically, YOLOV4) to recognize a given word. If the word was recognized, accurate text generation for that word was achieved, otherwise no text was generated.

The system has the ability to identify Bengali words from video recordings of BdSL. The system can detect when a sign is employed to represent an entire word.

The scope of the study is restricted to the categorization of only 34 words in BdSL. The study lacks the ability to produce whole sentences.

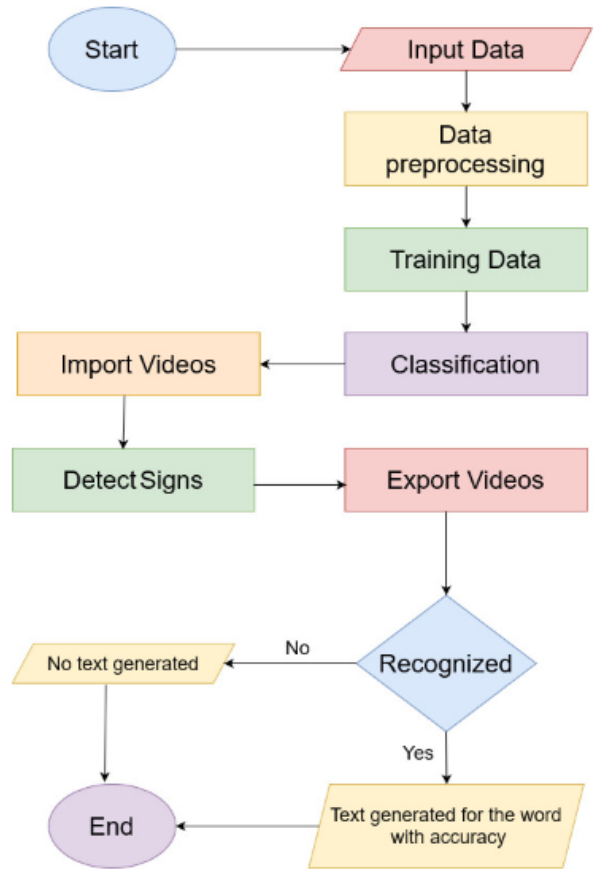

Figure 3.11: Overview of the system[10]

## 3.6   Sentence-Level Translation

Talukder et al.[11] created a dataset of 12.5k images divided into 49 different categories, 36 of which related to Bangla alphabets (including 6 Bangla vowels and 30 Bangla consonants), 10 to Bangla digits, and the remaining 3 to newly proposed signs for the formation of compound characters, spaces, and punctuation (specifically, dari or vertical bar). Their method entailed using YOLOv4 for object identification and word creation with a rate of 30 fps, based on the assumption that an individual would give one sign every second. When the system found two or more distinct signs in a one-second period, all of the signs were printed; however, if the system found similar signs in a series of frames, it would delay generating the word for 30 frames. Sentence and speech generation were also included in the study, with the latter being carried out utilizing the application programming interface (API) of Google Text to Speech (gTTS)

Ability to generate complete Bengali sentences directly from video input. This capability enables the system to provide a more comprehensive understanding of the signed message. The system is capable of generating output in both Bengali text and speech formats, which can cater to the preferences of different users.

The system is capable of generating output in both Bengali text and speech formats, which can cater to the preferences of different users. While Bengali language has 49 alphabets, the sign language only has 36, which creates a limitation in generating some common words in the language. The current approach for generating words follows alphabetical order, but this may not always align with natural usage patterns. While the paper provides a sign for the punctuation mark "dari" (|), there are no signs available for other punctuation marks. This could hinder proper sentence generation in Bengali sign language. The lack of decimal point signs makes it impossible to form floating point numbers in the language.

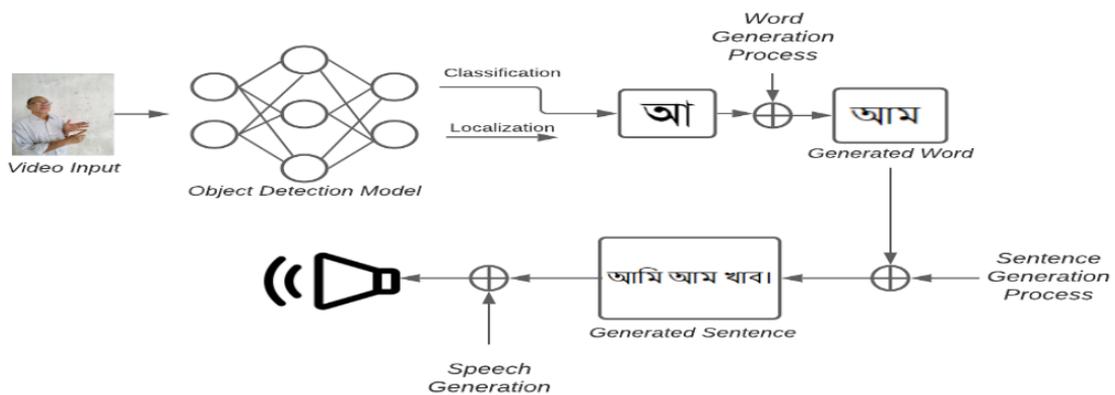

Figure 3.12: Overview of the system[11]

There is still a huge research vacuum in BdSL translation, because no previous work has specifically addressed the identification of continuous sign language, which would allow for the construction of full phrases. This offers a big chance for the field's further investigation and development.

# Chapter 4

# Research Methodology

Single-modal data in sign language recognition has inherent limitations, such as RGB images being susceptible to lighting and angle variations, and skeleton data lacking facial details. To overcome these drawbacks and enhance the representation of sign language, effective multimodal data fusion techniques are employed. By combining different modal data, the system leverages their strengths and specific attributes, resulting in robust sign language representations. Despite the potential benefits, currently, there are only a limited number of algorithms for continuous SLR utilize fusion of multimodal data strategies. In this research, the fusion of skeleton and RGB data is adopted, capitalizing on the resilience of skeletal data and the depth of information it provides of RGB data, thereby advancing the system's performance and accuracy.

In our method, we fuse two stream encoding process to encode the sign language videos, $V \in \mathbb{R}^{T \times N \times C}$. In the first stream, i3d features, $\mathbf{E_t} \in \mathbb{R}^{\mathcal{T} \times C}$, are extracted using an i3d network. Then, this feature is positionally encoded before being fed into a transformer encoder that is formed with several transformer encoder blocks,$N_{TE}$. The ultimate output encoding of this stream is denoted as $\mathbf{Z_t} \in \mathbb{R}^{\mathcal{T} \times C}$. In the second stream, we extract keypoint features using the Mediapipe algorithm. This process establishes a spatio-temporal graph structure, $G = (V, E, A)$, incorporating $K$ joints and $T$ frames. Then, the keypoint data is passed through a STGCN-LSTM Encoder that is formed by multiple STGCN blocks, $N_{STGCN}$, followed by a LSTM layer. This stream finally gives an encoding represented as $\mathbf{Y} \in \mathbb{R}^{\mathcal{T} \times N \times (C_{out})}$. This two stream is fused in the next step to pass the final encoding to a transformer decoder that is formed with multiple transformer decoder blocks, $N_{TD}$. The preprocessed text is also passed to the transformer decoder. The decoder generates tokenized output which is detokenized in the post processing step to finally generate the predicted spoken language text, $S \in \mathbb{R}^N$.

## 4.1 Joint Encoding

### 4.1.1 Video Encoding

We employ an I3D network[74] over 16-frame video clips, $F_{\text{I3D}}(V_{t:t+15})$, to give an embedding, $\psi_v$. This number of frame is taken because of its effectiveness in sign language recognition methods[29, 62]. Then, we temporarily aggregate the $\psi_v$ to a vector of constant size. Finally, the vector is transformed into the $C$ dimensional embedding space.

$$E_t = P(F_{\text{temporal}}(F_{\text{I3D}}(V_{t:t+15})))  \tag{4.1}$$

Here, $E_t \in \mathbb{R}^C$ is the feature embedding of the video, $F_{\text{temporal}(.)}$ is the temporal aggregation function and $P(.)$ is the projection operation. As for translating sign language, it is crucial to have information about the placement of sign gestures within the entire video sequence. But transformer network do not have access to positional information. To overcome this limitation, we introduce temporal ordering information to the embedded representations of the video using positional encoding $E_{pos}(.)$.

$$\hat{E}_t = E_t + E_{pos}(t)  \tag{4.2}$$

We proceed to train a transformer encoder model using the positional encoded embedding of the video frames, $\hat{E}_{1:T}$. Initially, the embedding of the video frames are passed to a self-attention layer, which is responsible for learning the contextual connections among the embedding of the video frames. Then, the results of the self-attention layers are forwarded through a non-linear point-wise feed forward layer. Throughout the process, residual connections and normalization are applied for effective training. The process can be formulated using the following equation, where $Z_t$ is the spatio-temporal representation of the frame $f_t$ at time step $t$, given the embedding of all the frames of the video, $\hat{E}_{1:T}$. Multiple encoder blocks are stacked to extract the features.

$$Z_t = \text{TransformerEncoder}(\hat{E}_t | \hat{E}_{1:T}) \quad (4.3)$$

## 4.1.2   Keypoint Encoding

We employ Mediapipe[68] algorithm for extracting keypoint feature. The MediaPipe Holistic Landmarker task allows for the integration of the pose, face, and hand landmark components, creating a comprehensive landmarker for the human body. This feature is ideal for the analysis of full-body gestures, poses, and actions, utilizing a machine learning model that operates on a continuous image stream. In real-time, this task delivers a combined total of 543 landmarks, consisting of 33 pose landmarks, 468 face landmarks, and 21 hand landmarks per hand. The assignment furnishes body pose landmarks in both image-based coordinates and three-dimensional global coordinates.

In the design of our architecture, we have exclusively employed the elements associated with pose landmarks. These pose landmarks are extracted using the MediaPipe Holistic Landmarker Model, which tracks 33 key points on the body, approximating the positions of the following body parts:

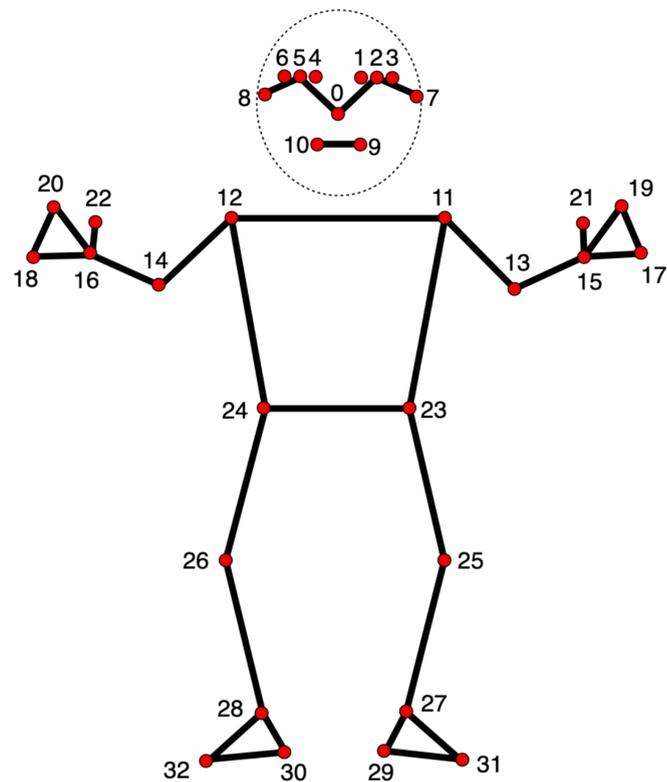

Figure 4.1: Pose Landmarker

0 - nose

1 - inner part of the left eye

2 - left eye

3 - outer part of the left eye

4 - inner part of the right eye

5 - right eye

6 - outer part of the right eye

7 - left ear

8 - right ear

9 - left side of the mouth

10 - right side of the mouth

11 - left shoulder

12 - right shoulder

13 - left elbow

14 - right elbow

15 - left wrist

16 - right wrist

17 - left pinky finger

18 - right pinky finger

19 - left index finger

20 - right index finger

21 - left thumb

22 - right thumb

23 - left hip

24 - right hip

25 - left knee

26 - right knee

27 - left ankle

28 - right ankle

29 - left heel

30 - right heel

31 - index finger of the left foot

32 - index finger of the right foot

Figure 4.2: Pose Landmark Keypoints

Then, we construct undirected spatio-temporal graph structure $G = (V, E, A)$ with $K$ joints and $T$ frames, comprising both intra-body and frame-by-frame connections. The set of vertices $V = \{V_{ti} \,|\, t = 1, \ldots, T, \, i = 1, \ldots, K\}$ includes all the joints $J_i$ in the sequence of frames. The feature vectors of each vertex $V_{ti}$ will be 3D coordinate vectors represented by $M_g(V_{ti}) = \{x, y, z\}$, where $x$, $y$, and $z$ are coordinates along each axis in a 3D coordinate space. The joints are connected with edges according to the connectivity of the human skeletal structure for a frame, and each of the joints is connected to itself through consecutive frames. $A \in \mathbb{R}^{N \times N}$ is the adjacency matrix of the graph $G$. To facilitate spatial configuration partitioning, we break down matrix $A$ into multiple matrices denoted as $A_k$, such that $\sum_k A_k = A + I$. $A_k$ represents the adjacency matrix of hop $k$. $A_0 = I$ and $A_1 = A$ were set as the initial conditions, where $I$ represents the identity matrix.

We input the keypoint data of a video, denoted as $V \in \mathbb{R}^{T \times N \times C}$, into the STGCN block to capture features from the body joints. Initially, we apply temporal convolution to the input keypoint sequences utilizing the kernel denoted as $\Gamma^u$. Subsequently, the input and temporal features are concatenated to extract spatial features.

$$\mathbf{Z} = \mathbf{X} \oplus (\Gamma^u \otimes \boldsymbol{V}) \tag{4.4}$$

Here, processed video is represented by $\mathbf{Z} \in \mathbb{R}^{\mathcal{T} \times N \times (C + C^u)}$. $C^\mu$ is the number of filter employed in temporal convolution, and the symbols $\otimes$ and $\oplus$ signify the temporal convolution and concatenation operations, respectively. Graph convolution is applied to $\mathbf{Z}$, incorporating the $k$th hop adjacency matrix denoted as $\mathcal{A}_k$ and utilizing the update rule of Graph Convolutional Networks (GCN)[26] to extract spatial features.

$$\mathbf{G}_k = \sigma(\tilde{\mathcal{A}}_k \mathbf{Z} \mathbf{W}_k) \tag{4.5}$$

Here, $\tilde{\mathcal{A}}_k = D_k^{-\frac{1}{2}}(\mathcal{A}_k + I)D_k^{-\frac{1}{2}}$, where $D$ represents a diagonal degree matrix, $\mathbf{W}_k$ is a weight matrix that can be learned, and $\sigma$ denotes a non-linear activation function. A linear transformation is done on the feature space using this equation, followed by the aggregation of neighboring information utilizing the normalized adjacency matrix. Then, three layers of Temporal Convolution Layers (TCNs) employing identical padding and kernel $\Gamma_1^l$, $\Gamma_2^l$ and $\Gamma_3^l$ respectively were implemented to obtain temporal features at various levels. We combined both advanced and foundational level features to identify movement patterns while performing a gesture in sign language at various level of complexity. The operations can be described as $m^1 = \Gamma_1^l \otimes \mathbf{G}_k$, $m^2 = \Gamma_2^l \otimes m^1$, $m^3 = \Gamma_3^l \otimes m^2$ and $m_k = m^1 \oplus m^2 \oplus m^3$, where features extracted from the $k$th hop is represented by $m_k$, considering both spatial and temporal aspects. Finally, output from each hop is concatenated.

$$\mathbf{Y} = m_1 \oplus m_2 \oplus \ldots \oplus m_k \tag{4.6}$$

The result of the STGCN block is denoted as $Y \in \mathbb{R}^{T \times N \times C_{\text{out}}}$. Here, $C_{\text{out}} = \sum_{i=1}^{k} C_l^i$ refers to a tensors of three dimensions.In our method, we stack multiple STGCN blocks to extract complex spatio-temporal features.

A single STGCN block is depicted below:

```
==================================================================
Layer (type:depth-idx)        Shape of Output         Parameters #
==================================================================
 Input:                       [-1, 3, 33, 60]         --
 Conv2d: 1-1                  [-1, 64, 33, 60]        1,792
 ReLU:                        [-1, 64, 33, 60]        --
 Concat:                      [-1, 67, 33, 60]        --
 Conv2d: 1-2                  [-1, 64, 33, 60]        4,352
 ReLU:                        [-1, 64, 33, 60]        --
 Conv2d: 1-3                  [-1, 64, 33, 60]        4,352
 ReLU:                        [-1, 64, 33, 60]        --
 Concat:                      [-1, 128, 33, 60]       --
 Conv2d: 1-4                  [-1, 16, 33, 60]        18,448
 ReLU:                        [-1, 16, 33, 60]        --
 Dropout2d: 1-5               [-1, 16, 33, 60]        --
 Conv2d: 1-6                  [-1, 16, 33, 60]        3,856
 ReLU:                        [-1, 16, 33, 60]        --
 Dropout2d: 1-7               [-1, 16, 33, 60]        --
 Conv2d: 1-8                  [-1, 16, 33, 60]        4,880
 ReLU:                        [-1, 16, 33, 60]        --
 Dropout2d: 1-9               [-1, 16, 33, 60]        --
 Concat:                      [-1, 48, 33, 60]        --
==================================================================
Total parameters: 37,680
Trainable parameters: 37,680
Non-trainable parameters: 0
Total multiplication-additions (M): 74.13
==================================================================
Input size (MB): 0.02
Forward/backward pass size (MB): 3.63
Parameters size (MB): 0.14
Estimated Total Size (MB): 3.79
```

After that, we pass the output to the last STGCN block, $Y^l$, to an LSTM block. The LSTM block captures the sequential dependencies in spatio-temporal feature vectors. For this, $Y^l$ were reshaped to $Y^r \in \mathbb{R}^{T \times NC_{out}}$, which is the input to the LSTM. The input given to the LSTM is denoted as $\mathbf{Y}_r^t \in \mathbb{R}^{NC_{out}}$, the $t$th sequence of $Y_r$. LSTM involves the following operations:

$$f^t = \sigma_g(\mathbf{W}_f \mathbf{Y}_r^t + \mathbf{U}_f h^{t-1} + b_f)$$
$$i^t = \sigma_g(\mathbf{W}_i \mathbf{Y}_r^t + \mathbf{U}_i h^{t-1} + b_i)$$
$$o^t = \sigma_g(\mathbf{W}_o \mathbf{Y}_r^t + \mathbf{U}_o h^{t-1} + b_o)$$
$$\tilde{C}^t = \tanh(\mathbf{W}_c \mathbf{Y}_r^t + \mathbf{U}_c h^{t-1} + b_c)$$

where, the matrices $\mathbf{W}_f, \mathbf{W}_i, \mathbf{W}_o$, and $\mathbf{W}_c$ with dimensions $\mathbb{R}^{C \times NC_{out}}$ represent the weight matrices responsible for mapping inputs to the three gates and input cell state. Similarly, $\mathbf{U}_f, \mathbf{U}_i, \mathbf{U}_o$, and $\mathbf{U}_c$ with dimensions $\mathbb{R}^{C \times C}$ denote the weight matrices associated with the previous hidden state. Additionally, the vectors $b_f, b_i, b_o$, and $b_c$ with dimensions $\mathbb{R}^C$ represent bias vectors for the four components. Here $C$ is the output space dimensionality of LSTM. The final cell state and the hidden states are calculated as follows:

$$C^t = f^t \odot \tilde{C}^{t-1} + i_t \odot \tilde{C}^t$$
$$h^t = o^t \odot \tanh(C^t)$$

We used a LSTM layer after the STGCN layers because of its ability to take into account of the change of spatio-temporal features along temporal dimension. LSTM is very helpful in processing of variable length sequences. In sign language, same sign done by different signers can have different length because of doing it in different pace. Thats why, LSTM is a better fit in this sequence problem solution. Finally from this STGCN-LSTM architecture, the resulting dimensions would be $H^{T \times N \times C}$ , where $C$ is the output space dimensionality of the LSTM.

### 4.1.3  Two-Stream Fusion Module

The fusion process integrates the I3D feature obtained from the transformer encoder, denoted as $\psi_v$, and the encoding from the STGCN-LSTM architecture, denoted as $H$. The fused encoding, denoted as Fused Encoding, is mathematically represented as:

$$\text{Fused Encoding} = \psi_v + H \tag{4.7}$$

The summation operation encapsulates information from both sources, creating a comprehensive representation that effectively captures both spatial and temporal aspects of the input video.

## 4.2   Text Encoding

For text encoding, we used Sentencepiece tokenizer[75], which segments the text into sub-word units. When we split longer sentence to sub-word units, that enable us to learn a better representation of phonetic variants and compound words. This approach is beneficial for the acquisition of less common words and addresses issues related to out of vocabulary words. After tokenization, positional information is introduced for having temporal order information of the text. Following is the formulation of this process, where $\hat{\mu} \in \mathbb{R}^C$ is the embedded representation of the text and $E_{pos}(.)$ is the positional encoding function for the text.

$$\hat{\mu} = \text{SubWordEmbedding}(w_n) + W_{pos}(n) \tag{4.8}$$

## 4.3   Decoding Output

At the beginning of the target spoken language sentence, we add special token $< bos >$. Then, position encoded word embedding is fed to a masked self-attention layer. Masking ensures that while gathering contextual information, each token only considers its preceding tokens. Marking is necessary as the decoder do not have access to future output token during inference. The combined video embedding and text embedding is passed to the decoder layer to learn the mapping between the reference and target sequence. The transformer decoder is trained to generate target words sequentially until it reaches the $< eos >$ token. The decoding process is formulated as:

$$H_{n+1} = \text{TransformerDecoder}(\hat{\mu}_u | \hat{\mu}_{1:n-1}, Z_{1:T}) \tag{4.9}$$

The overall probability of the sentence, $p(S|V)$ is calculated by multiplying the probabilities of individual words given their respective contexts, $H_u$. This can be formulated as:

$$p(S|V) = \prod_{u=1}^{n} p(w_u | H_u) \tag{4.10}$$

## 4.4   Loss Optimization

We employed Labeled Smoothed Cross Entropy (LSCE) loss for training our method. This loss is a modified version of cross entropy with the integration of label smoothing, which is a regularization technique that helps the model to overfit. LSCE loss encourages the model to generalize translation to unseen translation. LSCE loss can be formulated by the following equation, where $\mathbf{y}$ represents the target translation, and $\hat{\mathbf{y}}$ signifies the predicted translation, $V$ is the size of the target vocabulary, and $\alpha$ is the label smoothing parameter.:

$$\begin{aligned} LSCE(\mathbf{y}, \hat{\mathbf{y}}) = -\frac{1-\alpha}{V-1} \sum_{i \neq y} \log \left( \frac{\exp(\hat{y}_i)}{\sum_{j=1}^{V} \exp(\hat{y}_j)} \right) \\ -\alpha \log \left( \frac{\exp(\hat{y}_y)}{\sum_{j=1}^{V} \exp(\hat{y}_j)} \right) \end{aligned} \tag{4.11}$$

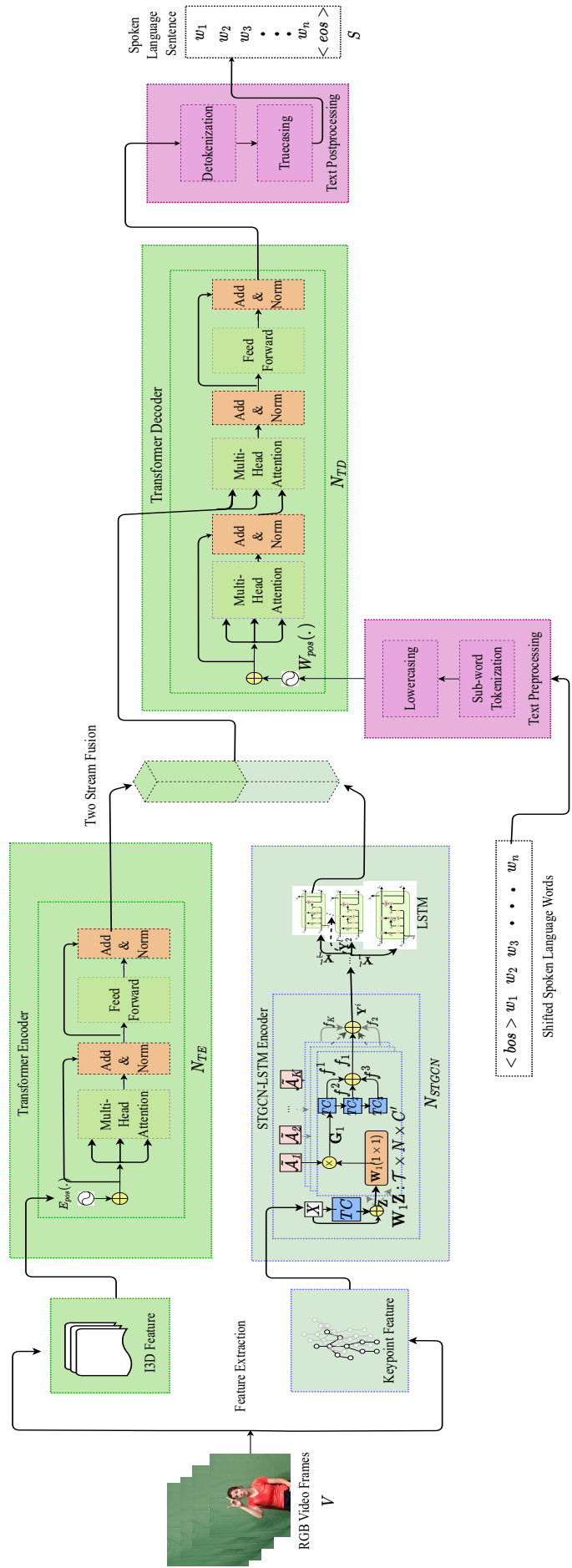

Figure 4.3: Methodology Overview [drawn in draw.io]

### 4.4.1   Implementation Details

In the training phase of our model, we employed a batch size of 32 and conducted 250 epochs, with each epoch requiring approximately 2.5 minutes for processing on a single NVIDIA GeForce RTX 3090 GPU. To ensure the accuracy and reliability of our model, we conducted validation every 2 epochs, enabling us to assess its performance and make necessary adjustments.

During the inference stage, we adopted established techniques commonly utilized in machine translation. Specifically, during the generation process of the text, by randomly choosing from the probability distribution, the decoder anticipates the next token., which is dependent on the tokens generated earlier in the sequence. Rather than solely choosing the anticipated outcome with the highest probability, we utilized the algorithm for making predictions of the beam search. This approach allowed us to generate multiple candidate sequences, with a chosen beam size of five, thereby enhancing the diversity and quality of the generated output.

Here are some of the hyperparameters that we used for training our model:

Table 4.1: Hyperparameters Used

| Hyperparameters | Values |
| --- | --- |
| Text Processing | Yes |
| Batch Size | 32 |
| Layers of Encoder-Decoder | 6-3 |
| Attention Heads | 4 |
| Learning Rate | 1e-3 |
| Feed-Forward Network Dimension | 1024 |
| Embedding Dimension | 256 |
| Learning Rate Scheduler | cosine |

## 4.5   Evaluation Protocol

### 4.5.1   BLUE

The BLUE-n[76] metric measures the degree of accuracy of machine translations compared to reference translations, employing n-grams as the basis of computation. An n-gram refers to a sequence of n words that appear consecutively. Common values of n that are frequently used include 1, 2, 3, and 4.

In computing the BLUE-n score, the accuracy of each n-gram in the translation is calculated by counting the occurrences of n-grams that appear in the reference translation. Furthermore, the precision is weighted by the length of the translation, as longer translations are more prone to having coincidental matches. The resulting weighted precision values are subsequently averaged across all n-grams to yield the final BLUE-n score.

The formula for computing the BLUE-n score is calculated by:

$$\text{BLEU-n} = \text{BP} \times \exp\left(\frac{1}{n}\sum_{i=1}^{n}\log(p_i)\right) \tag{4.12}$$

where BP is a brevity penalty that corrects for translations that are shorter than the reference, and $p_i$ is the precision for n-grams of length i.

The brevity penalty is computed as follows:

$$\text{BP} = \begin{cases} 1 & \text{if } \text{MT}_{\text{length}} > \text{Ref}_{\text{length}} \\ \exp\left(1 - \frac{\text{Ref}_{\text{length}}}{\text{MT}_{\text{length}}}\right) & \text{if } \text{MT}_{\text{length}} \leq \text{Ref}_{\text{length}} \end{cases} \tag{4.13}$$

where $\text{MT}_{\text{length}}$ represents the length of the machine translation, and $\text{Ref}_{\text{length}}$ denotes the length of the reference translation.

### 4.5.2   Reduced BLEU

Motivated by the research presented in reference [77], we comuted reducedBLEU (rBLEU). This metric involves the selective exclusion of specific words taking into account words from both the reference and the prediction prior to computing the BLEU score

A blacklist of words are created that are frequently encountered in the training data but have limited impact on the overall meaning of sentences. These words are excluded while calculatng reducedBLEU score. These words typically encompass articles, prepositions, pronouns, and other such linguistic elements. By implementing this approach, we aim to refine the evaluation of language generation models, emphasizing the importance of content-rich words while mitigating the influence of less significant linguistic components. This innovative method promises to provide a more nuanced assessment of text generation quality.

The identified blacklisted words for German language are top 100 most frequent words used in the dataset's vocabulary. Some of these words are:

"und", "im", "es", "der", "am", "in", "die", "auch", "grad", "bis", "den", "morgen", "nacht", "noch", "an", "heute", "regen", "sich", "wolken", "ein", "süden", "sonne", "norden", "ist", "wird", "für", "das", "schauer", "mit", "von", "westen", "aber", "dann", "sonst", "tag", "wind", "da", "osten", "wieder", "auf", "mal", "über", "teilweise", "gewitter", "aus", "vor", "zum", "oder", "nur", "wettervorhersage", "nun", "schnee", "etwas", "hier", "meist", "wetter", "abend", "regnet", "gibt", "nordwesten", "uns", "bleibt", "allem", "nach", "minus", "zu", "bei", "freitag", "sind", "teil", "richtung", "weht", "dabei", "freundlich", "dem", "donnerstag", "südosten", "sonntag", "samstag", "dort", "luft", "werden", "trocken", "nebel", "nordsee", "temperaturen", "mittwoch", "südwesten", "kann", "mäßig", "schon", "einzelne", "nordosten", "montag", "so", "wir", "schwach", "mitte", "fällt", "alpen"

Figure 4.4: German Blacklisted Words List

To understand the words distribution in the German dataset's dictionary, top 50 common words are in the plotted in histogram.

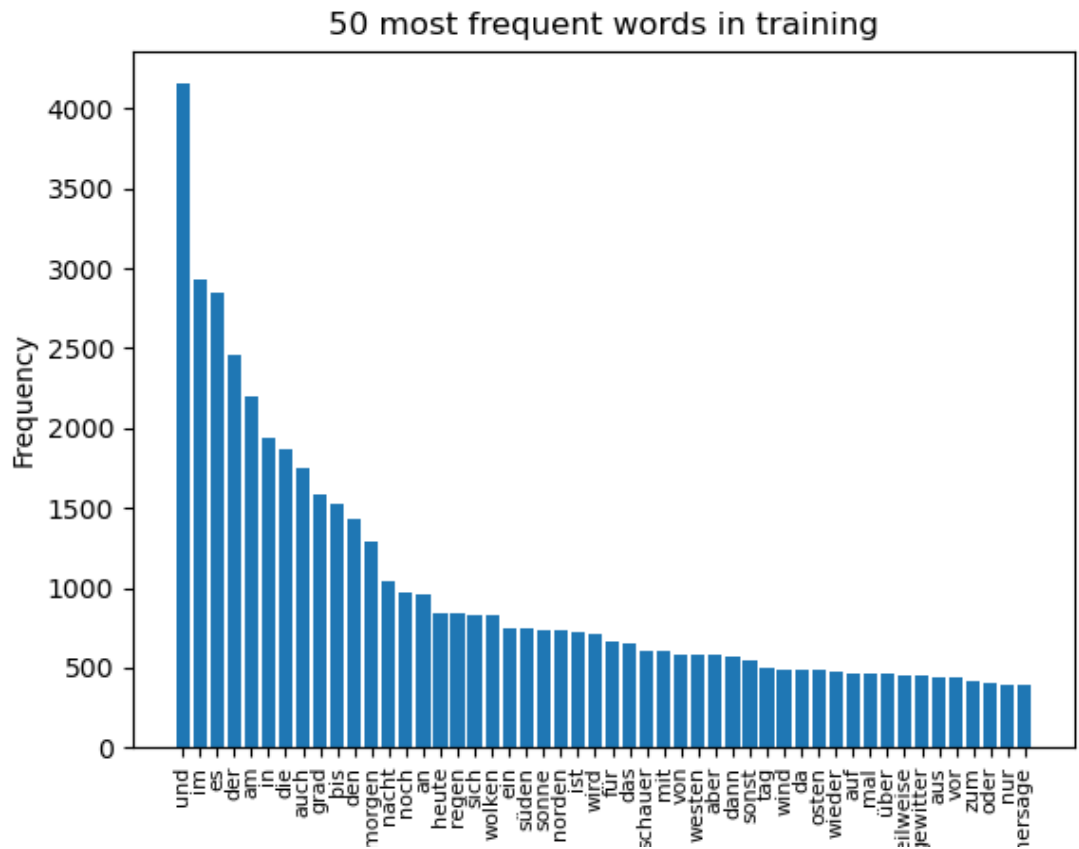

Figure 4.5: German Blacklisted Words Frequency

The identified blacklisted words for English language are top 290 most frequent words used in the dataset's vocabulary. Some of these words are:

'the', 'to', 'and', 'you', 'a', 'of', 'that', 'is', 'it', 'in', 'your', 'going', 'this', 'so', 'on', 'i', 'have', 'can', 'with', 'are', 'just', 'we', 'be', 'want', 'for', 'do', 'if', 'or', 'up', "it's", 'as', "you're", 'like', 'get', 'one', "i'm", 'what', 'now', 'little', 'out', 'then', 'here', 'make', 'go', 'but', 'not', 'about', "we're", 'when', 'they', 'all', 'there', 'my', 'will', 'at', 'how', 'some', 'right', 'back', 'really', 'use', 'very', 'down', "don't", "that's", 'because', 'our', 'way', 'from', 'them', 'take', 'more', 'put', 'an', 'good', 'into', 'see', 'sure', 'bit', 'also', 'need', 'these', 'would', 'know', 'off', 'lot', 'over', 'two', 'kind', 'which', 'thing', 'time', 'other', 'where', 'look', 'got', 'first', 'again', 'any', 'keep', 'actually', 'nice', 'has', 'side', 'different', 'come', 'around', 'start', 'things', 'hair', "there's", 'through', 'ball', 'by', 'her', 'much', 'well', 'their', 'people', 'three', 'hand', 'something', 'those', 'important', 'talk', 'top', 'next', 'water', 'work', "they're", 'even', 'most', 'doing', 'once', 'using', 'he', 'same', 'show', 'step', 'might', 'okay', 'body', 'give', 'me', 'find', 'too', 'another', 'always', "you'll", 'could', 'try', 'before', 'left', 'line', "you've", 'great', 'say', 'than', 'was', 'cut', 'ahead', 'hi', 'think', 'color', 'part', 'help', "let's", 'pretty', 'place', 'today', 'should', 'turn', "we'll", 'may', 'let', 'maybe', 'add', 'head', 'type', "i've", 'no', 'front', 'his', 'skin', 'pull', 'forward', 'foot', 'long', 'set', 'done', 'piece', 'four', 'end', 'bring', 'able', 'hands', "doesn't", 'getting', 'many', 'called', 'move', 'every', 'feel', 'big', 'feet', 'together', 'only', 'open', 'best', 'straight', 'hold', 'leg', 'point', "we've", 'inside', 'each', 'still', 'sometimes', 'area', 'does', 'been', 'him', 'paper', 'basically', 'bottom', 'remember', 'used', 'after', 'probably', 'name', 'usually', 'away', 'comes', 'times', 'making', 'tell', 'ok', "i'll", 'goes', 'yourself', 'five', 'small', 'coming', 'talking', 'had', 'ready', 'am', 'half', 'play', 'she', 'oil', 'were', 'anything', 'looking', 'face', 'glass', 'working', 'lift', 'day', 'everything', 'hit', 'brush', 'easy', 'light', 'clean', 'dog', 'while', 'its', 'car', 'better', 'who', "he's", 'enough', 'white', 'position', 'being', 'new', 'person', 'looks', 'clip', 'high', 'course', 'made', 'throw', 'between', 'couple', 'buy', 'basic', 'dry', 'practice'

Figure 4.6: English Blacklisted Words List

To understand the words distribution in the English dataset's dictionary, top 50 common words are in the plotted in histogram.

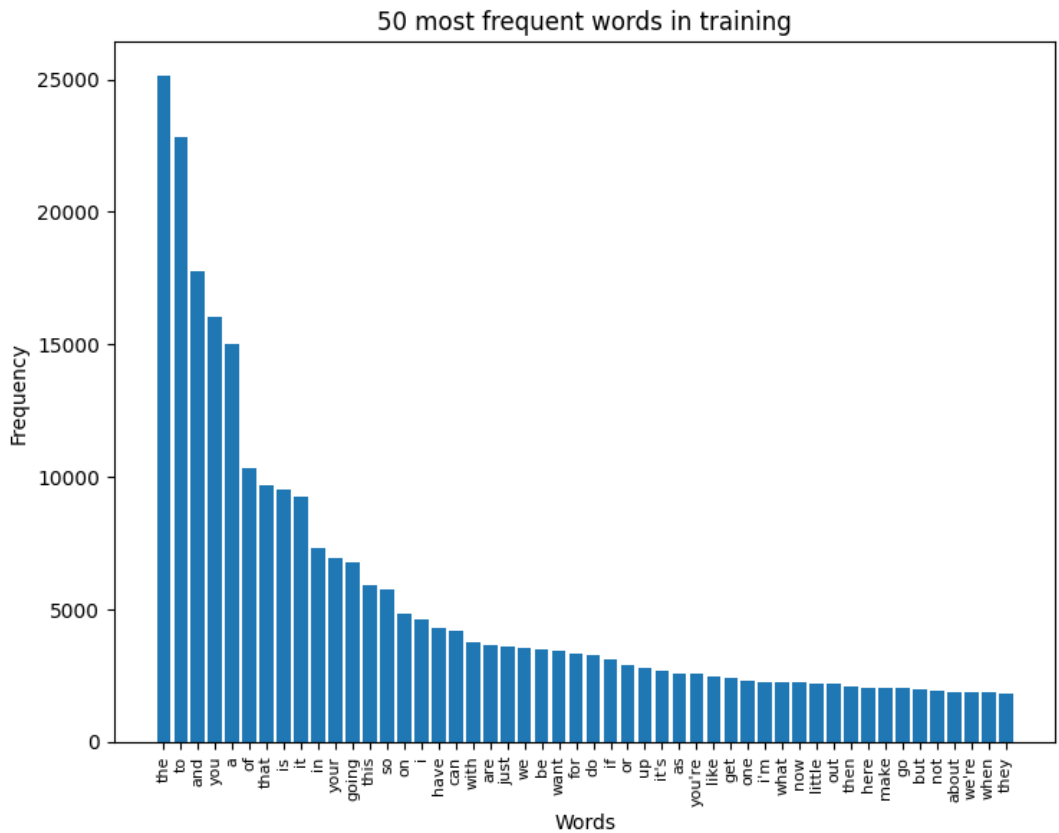

Figure 4.7: English Blacklisted Words Frequency

Similarly to identify the blacklisted words in the Bengali language, we conducted an analysis and pinpointed the 300 most commonly occurring words within our dataset's vocabulary. These blacklisted words are mentioned here.

না, ও, আমি, করে, না।, এই, তিনি, আমার, তার, থেকে, করতে, আর, সে, এ, জন্য, কথা, হয়, হবে, এক, করা, আমাদের, কিন্তু, তো, কী, এটা, কি, তুমি, সঙ্গে, আমরা, এবং, আছে, বলেন, একটা, হয়, কাজ, নিয়ে, যে, হয়েছে, একটি, বলে, এখন, মধ্যে, এর, কোনো, কিছু, করেন, ছিল, হবে।, অনেক, মনে, তোমার, সময়, তা, সাথে, তাদের, ভালো, সব, আছে।, দুই, খুব, তাঁর, দেশের, কেন, হয়।, একজন, পর, আপনি, দেখা, করার, তারা, করে।, নেই।, হচ্ছে, , ঠিক, আমাকে, তাই, সেই, তাকে, দিয়ে, আগে, শুরু, এখানে, ওই, তবে, কাছে, বা, বিডিনিউজ, বলা, যা, শেষ, যায়, বিভিন্ন, আবার, সবাই, টাকা, বড়, কোন, বছর, প্রথম, টোয়েন্টিফোর, কেউ, দিতে, একটু, ছিলেন, আপনার, বাবা, করেন।, হলো, ঢাকা, তখন, বাংলাদেশ, ছিল।, বাংলাদেশের, জানান, করেছেন, নতুন, মানুষের, গত, মানুষ, চলে, চাই, বেশি, যখন, দিন, আরো, হয়, আওয়ামী, যেতে, হলে, যেন, পরে, হয়, এসব, প্রতি, যাবে, এমন, মা, কোথায়, গেছে।, ধরে, আজ, সেটা, নেই, প্রধান, সরকার, গিয়ে, হতে, করবে, রয়েছে, আগামী, করছে, বিষয়ে, বড়, ছোট, দেখে, দিকে, নির্বাচন, তথ্য, বলতে, হয়েছে।, যায়।, পুলিশ, একই, সরকারের, পারে, করবেন, যদি, বছরের, ইসলাম, উপর, হাজার, ধরনের, এটি, খালেদা, শুধু, দেখতে, নয়।, হয়।, বের, খেলা, পর্যন্ত, যোগাযোগ, পারে।, থাকে, দেশে, দলের, মতো, কর।, কারণে, ছেলে, ভাই, ওকে, বেশ, নিয়ে, বন্ধ, গেছে, নামে, শিক্ষা, ব্যবস্থা, পারি, প্রায়, অন্য, প্রধানমন্ত্রী, হাত, হিসেবে, লাখ, সাধারণ, হোসেন, এস, মারা, যারা, নাম, দেওয়া, জানি, হাতে, সকালে, দিয়ে, রাখতে, আরও, তৈরি, নয়, প্রকাশ, দুটি, শেখ, বললেন, করেছে, সম্পাদক, এবার, আপনাকে, তিন, তোমাকে, কে, ছাত্র, করো।, তাঁকে, যাও।, সালের, চেয়ে, করি, এত, বসে, লীগ, দাবি, ছবি, আহত, শিশু, প্রশ্ন, কোটি, চৌধুরী, এখনো, রাত, রহমান, নেওয়া, মঙ্গলবার, যায়, বিশেষ, নিজেদের, রাজা, এতে, নিজের, পাওয়া, সহায়তা, ছিলেন।, নিচে, পড়ে, যাবে।, অংশ, পানি, এরপর, থাকতে, বিশ্বাস, শুনে, করলে, নাই।, স্যার।, চার, টি, ওখানে, তারপর, একবার, দাও।, মানে, মাঝে, ফলে, ফেলে, দল, বক্তব্য, গেলে, আসলে, সম্পর্কে, সরকারি, থেকেই, বিএনপি, জিয়া, জীবন, সুন্দর, সেখানে, আবে, ফিরে

Figure 4.8: Bangla Blacklisted Words List

To understand the words distribution in the Bangla dataset's dictionary, top 50 common words are in the plotted in histogram.

Figure 4.9: Bangla Blacklisted Words Frequency

## 4.6    Dataset Description

We experimented our method on three publicly available datasets: RWTH-PHOENIX-2014T[4], CSL-Daily[12], and How2Sign[13].

### 4.6.1    RWTH-PHOENIX-2014T

RWTH-PHOENIX-2014T dataset[3] comprises 11 hours of continuous German Sign Language (DGS) recordings performed by 9 individuals. It contains both transcription and gloss data, but it should be mentioned that it is a single-view dataset, implying that the videos were captured from a single perspective. Furthermore, It is important to mention that information related to speech and body posture is noteworthy that are absent from the dataset. Despite these limitations, the dataset offers a valuable resource for researchers and developers interested in studying German Sign Language, enabling the exploration of the language's intricacies and the development of innovative sign language technologies.

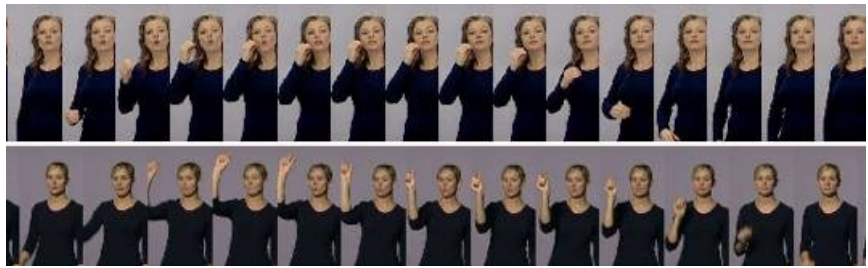

Figure 4.10: RWTH-Phoenix-2014T Dataset[3]

### 4.6.2   CSL-Daily

CSL-Daily[12] is a dataset on Chienese Sign Language (CSL). The dataset contains sign language videos, text and gloss annotations on Daily Life events, such as domestic life, healthcare, educational experiences, banking services, retail activities, interpersonal interactions, and similar aspects. The dataset consists of 25000 samples that are done by 50 signers. Along with chinese sign language videos, transcription and gloss are available for the videos.

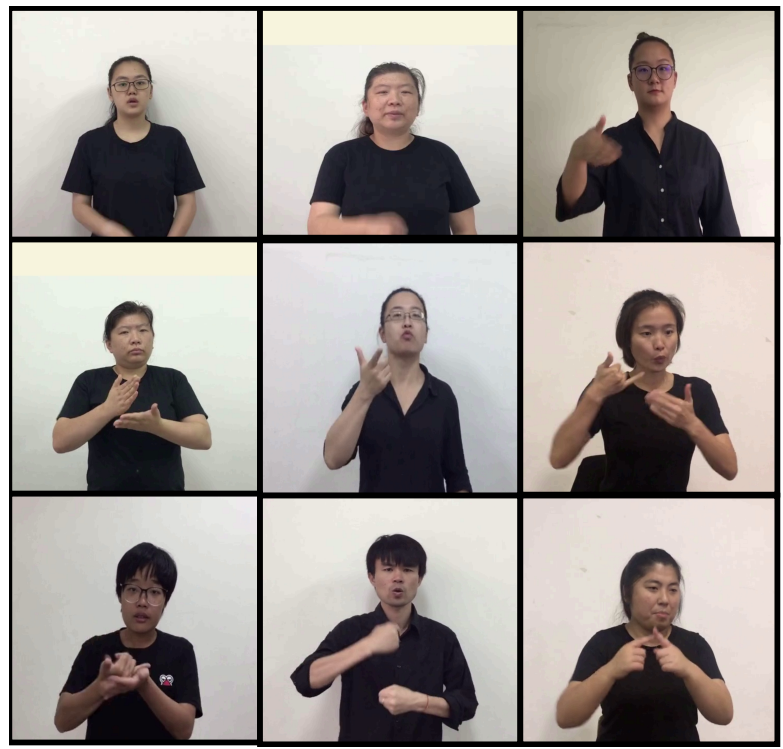

Figure 4.11: Signers of CSL-Daily Dataset[12]

### 4.6.3   How2Sign

The How2Sign dataset[13] is a extensive, multimodal, and encompasses multiple view collection specifically tailored for sign language translation and recognition tasks. With a corpus of over 80 hours of continuous American Sign Language (ASL) videos, the dataset offers a comprehensive resource. It encompasses various modalities, including speech, English transcripts, and depth information, enhancing the richness of the dataset. The dataset focuses on ASL and features a total of 11 signers. The videos in the dataset are continuous, allowing for a more natural representation of sign language communication. The instructional videos cover a variety of topics, enabling sign language users to demonstrate different signs while providing verbal explanations. Researchers in the field of sign language understanding and translation can utilize the How2Sign dataset to explore various aspects of sign language communication. It serves as a valuable resource, facilitating advancements in the making of robust and accurate models for sign language recognition and translation.

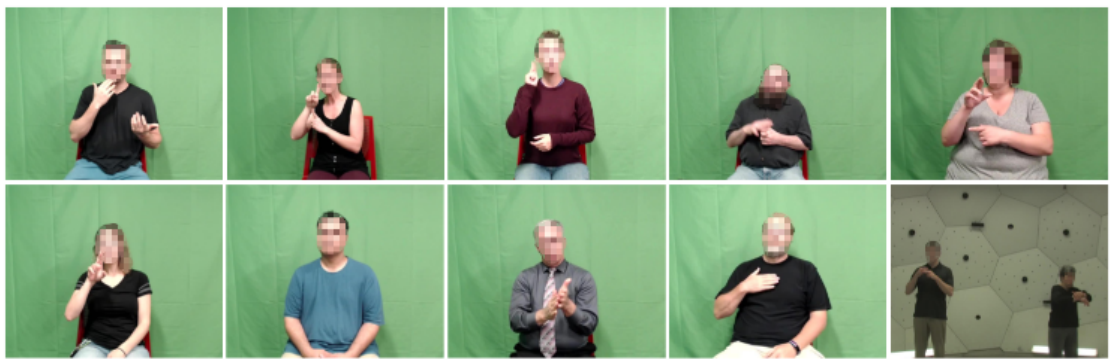

Figure 4.12: The How2Sign dataset comprises videos featuring 11 different signers. In the Green Screen Studio, the top row showcases signer 1- signer 5 from the left to right, while the bottom row features signers 8-11 in the same order. An additional figure on the bottom row displays signer 6 - signer 7, which were recorded in the Panoptic studio.[13]

### 4.6.4   BornilDB v1.0

BornilDB v1.0 dataset[14] comprises 45 hours of recordings of Bangla Sign Language (BdSL) executed by three individuals. It is important to note that this dataset is not a multiview dataset, as the videos have been filmed from a solitary viewpoint. Despite the inclusion of transcription data, it is worth noting that the dataset does not contain any gloss information. Furthermore, the dataset does not feature any pose or speech information.

Additionally, it is noteworthy to highlight the distinctive characteristics of the BornilDB v1.0 dataset in comparison to the other three datasets used in the study. While the other three datasets utilized a green screen as a background, BornilDB v1.0 features a dynamic background consisting of elements such as beds, clothes, windows, sofas, etc. Moreover, unlike the other datasets that maintained uniform video dimensions recorded with the same configuration cameras, this dataset used different devices with varying configurations, resulting in different video dimensions. These factors contribute to the dataset's increased complexity and make it more challenging compared to the others.

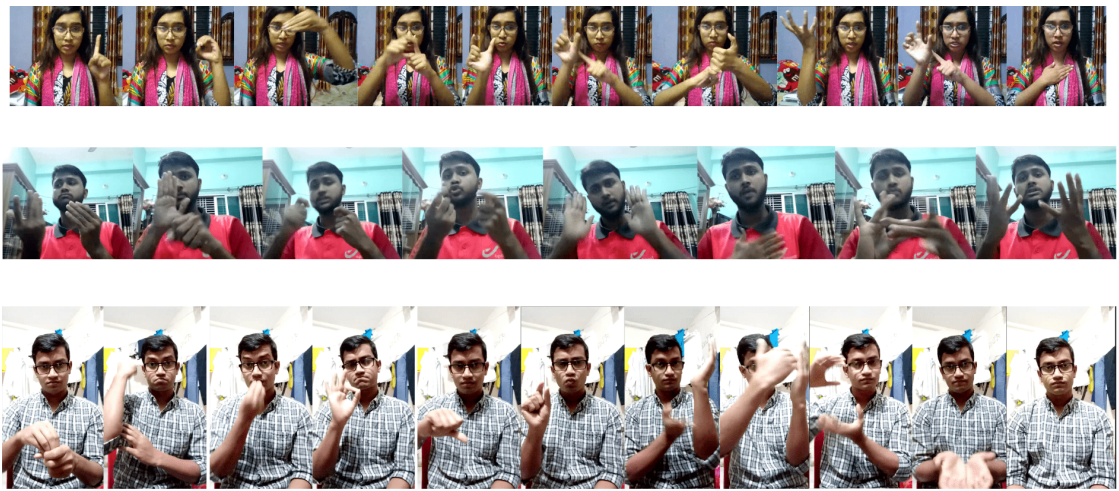

Figure 4.13: BornilDB v1.0 Dataset[14]

A comparative analysis of these datasets is presented in Table 4.2.

| Dataset | RWTH-PHOENIX-2014T | CSL-Daily | How2Sign | BornilDB v1.0 |
|---|---|---|---|---|
| Language | DGS | CSL | ASL | BdSL |
| Year | 2018 | 2015 | 2021 | 2023 |
| Language Level | Continuous | Continuous | Continuous | Continuous |
| Signers | 9 | 50 | 11 | 3 |
| Duration (h) | Train: 9.2 Val: 0.6 Test: 0.7 | Train: 20.62 Val: 1.24 Test: 1.41 | Train: 69.6 Val: 3.9 Test: 5.6 | 73 |
| Vocabulary (k) | Train: 2 Val: 0.9 Test: 1 | Train: 2 Val: 1.3 Test: 1.3 | Train: 15.6 Val: 3.2 Test: 3.6 | 25.57 |
| Total Samples | ✗ | 25,000 | ✗ | 21,154 |
| Domain | Weather Forecast | Daily Life | Instructional | |
| Multiview | ✗ | ✓ | ✓ | ✗ |
| Transcription | ✗ | ✓ | ✓ | ✗ |
| Gloss | ✓ | ✓ | ✓ | ✓ |
| Pose | ✓ | ✓ | ✓ | ✓ |

Table 4.2: Comparative Analysis of SLT Datasets Based on Language, Number of Signers, Video Duration (in hours), Vocabulary Size (in thousands), and Domain Attributes

# Chapter 5

# Result

## 5.1 Result on RWTH-PHOENIX-2014T Dataset

In Table 5.1, we present the comparative results, which include BLEU scores of our model in contrast to comparison models on the RWTH-PHOENIX-2014T dataset[3]. The results for Joint-SLT[4] were obtained from [78], where they reproduced the results in a gloss-free context, while for other models, we relied on the results originally reported in their respective papers. As illustrated in Table 5.1, our model shows a notable improvement in translation performance, surpassing all state-of-the-art models significantly.

| Methods | BLEU-1 | BLEU-2 | BLEU-3 | BLEU-4 |
|---|---|---|---|---|
| Conv2d-RNN[3] | 27.10 | 15.61 | 10.82 | 8.35 |
| + Luong Attn.[3] + [79] | 29.86 | 17.52 | 11.96 | 9.00 |
| + Bahdanau Attn. [3] + [80] | 32.24 | 19.03 | 12.83 | 9.58 |
| Joint-SLT [4] | 30.88 | 18.57 | 13.12 | 10.19 |
| Tokenization-SLT [46] | 37.22 | 23.88 | 17.08 | 13.25 |
| TSPNet-Sequential [5] | 35.65 | 22.80 | 16.60 | 12.97 |
| TSPNet-Joint [5] | 36.10 | 23.12 | 16.88 | 13.41 |
| GASLT [78] | 39.07 | 26.74 | 21.86 | 15.74 |
| Our Method | 41.19 | 30.89 | 24.23 | 19.75 |

Table 5.1: Performance Comparison of Gloss-Free Translation Methods in the RWTH-PHOENIX-2014T Dataset

Qualitative result of our method is reported here. In table 5.2, generated translation of RWTH-PHOENIX-2014T[3] is shown using our best performing model. as the reference and generated translation are in German, English translation is provided also for better understanding.

Table 5.2: Reference and Prediction with English Translation

| No. | Type | German (English Translation) |
|---|---|---|
| 1 | Reference: | guten abend liebe zuschauer |
| | | (good evening dear viewers) |
| | Prediction: | guten abend liebe zuschauer |
| | | (good evening dear viewers) |
| 2 | Reference: | ihnen noch einen schönen abend |
| | | (Have a nice evening) |
| | Prediction: | ihnen noch einen schönen abend |
| | | (Have a nice evening) |
| 3 | Reference: | hallo und guten abend |
| | | (Hello and good evening) |
| | Prediction: | hallo und guten abend |
| | | (Hello and good evening) |
| 4 | Reference: | und nun die wettervorhersage für morgen dienstag den fünfundzwanzigsten mai |
| | | (And now the weather forecast for tomorrow, Tuesday May 25th) |
| | Prediction: | und nun die wettervorhersage für morgen dienstag den fünfundzwanzigsten mai |
| | | (And now the weather forecast for tomorrow, Tuesday May 25th) |

## 5.2    Result on CSL-Daily Dataset

In the table, we provide a comparison of results, presenting the BLEU scores of our model in contrast to other models using the CSL-Daily dataset[12]. The table reveals that our model demonstrates a considerable enhancement in translation performance, outperforming all state-of-the-art models with significant superiority.

| Methods | BLEU-1 | BLEU-2 | BLEU-3 | BLEU-4 |
|---------|--------|--------|--------|--------|
| Joint-SLT [4] | 21.56 | 8.29 | 3.68 | 1.72 |
| TSPNet-Joint [5] | 17.09 | 8.98 | 5.07 | 2.97 |
| GASLT [78] | 19.90 | 9.94 | 5.98 | 4.07 |
| Our Method | 22.34 | 11.67 | 6.45 | 6.14 |

Table 5.3: Performance Comparison of Gloss-Free Translation Methods on the CSL-Daily Dataset

We present the qualitative outcomes of our approach next. Table 5.4 illustrates the translated output of CSL-Daily[12] employing our most effective model. Since the reference and generated translations are in Chinese, we have included English translations for clarity and improved comprehension.

The translation quality of the sentences is highly impressive, as evidenced by the remarkable accuracy in capturing the intended meaning. Some instances even showcase predicted sentences that closely mirror the reference sentences. Furthermore, even when there are slight variations in wording, the Chinese sign language translation effectively conveys a similar meaning. This not only highlights the precision of the translations but also underscores the versatility in expression achieved through the model's predictions.

Table 5.4: Reference and Prediction with English Translation

| No. | Type | Chinese (English Translation) |
|---|---|---|
| 1 | Reference: | 警察要检查你的身份证。 |
| | | (The police want to check your ID.) |
| | Prediction: | 警察要检查你的身份证。 |
| | | (The police want to check your ID.) |
| 2 | Reference: | 这个月我每天游泳。 |
| | | (I swim every day this month.) |
| | Prediction: | 这个月我每天游泳。 |
| | | (I swim every day this month.) |
| 3 | Reference: | 这条裤子怎么样? |
| | | (How do these pants look?) |
| | Prediction: | 这条裤子怎么样? |
| | | (How do these pants look?) |
| 4 | Reference: | 我自愿加入组织。 |
| | | (I volunteered to join the organization.) |
| | Prediction: | 我自愿参加班。 |
| | | (I volunteered for the class.) |
| 5 | Reference: | 今天天气有点冷。 |
| | | (The weather is a bit cold today.) |
| | Prediction: | 今天的天气很冷。 |
| | | (It's very cold today.) |
| 6 | Reference: | 那你现在为什么不去? |
| | | (Then why don't you go now?) |
| | Prediction: | 他今天为什么不去? |
| | | (Why doesn't he go today?) |
| 7 | Reference: | 今天星期几? |
| | | (What day is it today?) |
| | Prediction: | 今天几月几日? |
| | | (What day is today?) |

## 5.3   Result on How2Sign Dataset

The effectiveness of our model on the How2Sign dataset[13], compared to the only available baseline for this dataset, is illustrated in Tables 5.5 and 5.6. Our model outperforms this baseline in all the metrics that were used for evaluation.

|                | Validation | | | | |
|----------------|------------|--------|--------|--------|--------|
|                | rBLEU | BLEU-1 | BLEU-2 | BLEU-3 | BLEU-4 |
| slt_how2sign[6] | 2.79 | 35.2 | 20.62 | 13.25 | 8.89 |
| Our Method | 3.97 | 37.37 | 22.18 | 14.45 | 9.9 |

Table 5.5: Performance Comparison of Gloss-Free Translation Methods on the How2Sign Dataset - Validation

|                | Test | | | | |
|----------------|------|--------|--------|--------|--------|
|                | rBLEU | BLEU-1 | BLEU-2 | BLEU-3 | BLEU-4 |
| slt_how2sign[6] | 2.21 | 34.01 | 19.3 | 12.18 | 8.03 |
| Our Method | 2.96 | 34.91 | 19.98 | 12.72 | 8.53 |

Table 5.6: Performance Comparison of Gloss-Free Translation Methods on the How2Sign Dataset - Test

The qualitative performance of our model on the How2Sign Dataset is shown in Table 5.7.

The translation quality of the sentences stands out for its impressive accuracy in capturing the intended meaning. Some instances even reveal predicted sentences that closely align with the reference sentences, showcasing the model's precision. Additionally, in cases where there are subtle variations in wording, the translation into American Sign Language effectively communicates a parallel meaning. This not only underscores the precision of the translations but also accentuates the model's capacity to convey versatility in expression within the realm of American Sign Language.

Table 5.7: Reference and Prediction with English Translation

| No. | Type | Translation |
| --- | --- | --- |
| 1 | Reference: | one more time. |
|  | Prediction: | one more time. |
| 2 | Reference: | the other thing i did was i had a little bit of line out. |
|  | Prediction: | another thing i would do is i have a little bit of line out. |
| 3 | Reference: | that's going to lead to overeating. |
|  | Prediction: | that's going to lead over food. |
| 4 | Reference: | and a very natural enhanced lip. |
|  | Prediction: | and very natural lips. |
| 5 | Reference: | just enough to build up some chest strength. |
|  | Prediction: | enough to build some chest strength. |
| 6 | Reference: | going to walk on to our next step. |
|  | Prediction: | and we' re going to walk on our next step. |
| 7 | Reference: | one and two and three and four. |
|  | Prediction: | and one, two, three, four. |
| 8 | Reference: | keep your eyes positioned up. |
|  | Prediction: | so keep your eye position up. |
| 9 | Reference: | the same thing over here. |
|  | Prediction: | same on the other side. |
| 10 | Reference: | it looks something like this. |
|  | Prediction: | the same thing, it looks like. |

## 5.4    Result on BornilDB v1.0 Dataset

In the absence of an established baseline for the BornilDB v1.0 Dataset [14], we sought to assess the efficacy of our model by conducting a comparative analysis with existing methods. To provide a comprehensive evaluation, we replicated several approaches on the dataset. Subsequently, our model demonstrated superior performance in comparison to these methods. The comparative results are shown in Tables 5.8 and 5.9.

|                    | Validation | | | |
|--------------------|--------|--------|--------|--------|
|                    | BLEU-1 | BLEU-2 | BLEU-3 | BLEU-4 |
| slt_how2sign[6]    | 4.32   | 1.39   | 0.54   | 0.27   |
| Our Method         | 7.62   | 3.05   | 1.37   | 0.72   |

Table 5.8: Performance Comparison of Gloss-Free Translation Methods on the BornilDB v1.0 Dataset - Validation

|                    | Test | | | |
|--------------------|--------|--------|--------|--------|
|                    | BLEU-1 | BLEU-2 | BLEU-3 | BLEU-4 |
| slt_how2sign[6]    | 4.25   | 1.11   | 0.25   | 0.27   |
| Our Method         | 7.37   | 2.89   | 1.18   | 0.58   |

Table 5.9: Performance Comparison of Gloss-Free Translation Methods on the BornilDB v1.0 Dataset - Test

Table 5.10 displays the translated results of the BornilDB v1.[14] dataset using our most successful model. Given that the reference and generated translations are in Bangla, English translations have been provided for enhanced clarity and understanding.

The excellence of sentence translation becomes evident through its remarkable precision in capturing the intended meaning. In certain cases, the predicted sentences closely resemble the reference ones, underscoring the model's accuracy. Moreover, when slight variations in wording occur, the translation into Bangla Sign Language adeptly maintains a similar meaning. This not only emphasizes the accuracy of the translations but also highlights the model's ability to convey versatility in expression within the context of Bangla Sign Language.

Table 5.10: Reference and Prediction with English Translation

| No. | Type | Bangla (English Translation) |
|-----|------|------------------------------|
| 1 | Reference: | সব কিছু ঠিক আছে তো? |
|   |            | (Is everything okay?) |
|   | Prediction: | সব ঠিক আছে? |
|   |             | (Is everything fine?) |
| 2 | Reference: | না, আমার মনে হয় না। |
|   |            | (No, I don't think so.) |
|   | Prediction: | না, আমার মনে হয় নেই। |
|   |             | (No, I don't feel like it.) |
| 3 | Reference: | আমি ভালো, বেশ ভালো। |
|   |            | (I'm good, pretty good.) |
|   | Prediction: | আমি তো অনেক ভাল আছি। |
|   |             | (I am very well.) |
| 4 | Reference: | আমায় ছেড়ে দাও! |
|   |            | (Leave me alone!) |
|   | Prediction: | আমাকে ছেড়ে দাও। |
|   |             | (Let me go.) |
| 5 | Reference: | ও এসে গেছে! |
|   |            | (He has come!) |
|   | Prediction: | সে এসে গেল। |
|   |             | (He has come.) |
| 6 | Reference: | সে জানত না? |
|   |            | (Didn't he know?) |
|   | Prediction: | সে জানে না? |
|   |             | (Does he not know?) |

## 5.5   Ablation Study

To enhance the structure and pinpoint the model that performs optimally, we conducted our ablation investigation on the RWTH-PHOENIX-2014T[3] dataset. We tried out four distinct approaches. The effects of changing the number of STGCN layers, the number of LSTM layers, the number of transformer encoder and decoder layers, and the efficacy of various fusion strategies were among these.

### 5.5.1   Effects of Varying the Number of STGCN Layers

In our experiment, we investigate the effect of of having different number of STGCN layers. While increasing the number of STGCN layers let our method acquire better and complex representation, it also expose the method to a higher risk of overfitting. With this objective in mind, we train our method using one to six STGCN layers.

As it can be seen in Table 5.11, our methods ability to translate gets better with additional STGCN layers initially. But, as we continued to add more STGCN layers, the method overfits on the training data, leading to a performance decrease in the test set. For this reason, we use 3 layers STGCN in our best performing model.

| #Layer | Validation | | | | Test | | | |
|--------|-------|-------|-------|-------|-------|-------|-------|-------|
|        | BLEU1 | BLEU2 | BLEU3 | BLEU4 | BLEU1 | BLEU2 | BLEU3 | BLEU4 |
| 1 | 40.88 | 30.40 | 23.86 | 19.57 | 41.02 | 30.03 | 23.25 | 18.87 |
| 2 | 40.99 | 30.41 | 23.96 | 19.72 | 41.05 | 30.27 | 23.62 | 19.26 |
| 3 | 40.97 | 30.26 | 23.6  | 19.23 | 41.19 | 30.89 | 24.23 | 19.75 |
| 4 | 41.40 | 30.61 | 23.66 | 19.19 | 40.99 | 30.15 | 23.45 | 19.06 |
| 5 | 40.09 | 29.75 | 23.27 | 18.95 | 41.22 | 30.45 | 23.64 | 19.16 |

Table 5.11: BLEU4 Scores for Different STGCN Layer Configurations

### 5.5.2 Effects of Varying the Number of Transformer Encoder and De-coder Layers

In our next experiment, we explore the impact of employing varying number of encoder and decoder layers in the Transformer architecture to find the optimum ones. While increasing the number of layers of encoder and decoder, our method acquire better and complex representation, it also expose the method to a higher risk of overfitting. With this objective in mind, we train our method using different combination of encoder and decoder layers.

As it can be seen in Table 5.12, our methods ability to translate gets better with additional layers of encoder and decoder. Our model performes best with 6 layers of encoder layer and 3 layers of decoder layer.

| # Encoder Layer | # Decoder Layer | BLEU1 | BLEU2 | BLEU3 | BLEU4 | Dataset Split |
|---|---|---|---|---|---|---|
| 6 | 3 | 40.97 | 30.26 | 23.6 | 19.23 | Validation |
|   |   | 41.19 | 30.89 | 24.23 | 19.75 | Test |
| 2 | 2 | 39.15 | 28.43 | 21.91 | 17.7 | Validation |
|   |   | 40.45 | 29.55 | 22.78 | 18.27 | Test |
| 3 | 3 | 41.22 | 30.29 | 23.55 | 19.16 | Validation |
|   |   | 40.69 | 30.23 | 23.53 | 19.1 | Test |

Table 5.12: BLEU4 Scores for Different Encoder and Decoder Layer Configurations

### 5.5.3 Effects of Varying the Number of LSTM Layers

We explored the impact of different number of LSTM layers in the STGCN-LSTM architecture to find the optimum number that gives us the best performing architecture. While increasing the number of LSTM layers let our method acquire better and complex representation, it also expose the method to a higher risk of overfitting. With this objective in mind, we train our method using one to three LSTM layers.

As it can be seen in Table 5.13, our methods ability to translate gets better with additional LSTM layers initially. But, as we continued to add more LSTM layers, the method overfits on the training data, leading to a performance decrease in the test set. For this reason, we use 1 layers LSTM in our best performing model.

| #Layer | Validation | | | | Test | | | |
|---|---|---|---|---|---|---|---|---|
| | BLEU1 | BLEU2 | BLEU3 | BLEU4 | BLEU1 | BLEU2 | BLEU3 | BLEU4 |
| 1 | 40.97 | 30.26 | 23.6 | 19.23 | 41.19 | 30.89 | 24.23 | 19.75 |
| 2 | 41.19 | 30.72 | 23.98 | 19.50 | 41.17 | 30.55 | 23.75 | 19.31 |
| 3 | 40.87 | 30.19 | 23.32 | 18.84 | 41.58 | 30.59 | 23.52 | 18.89 |

Table 5.13: BLEU4 Scores for Different LSTM Layer Configurations

## 5.5.4 Effectiveness of Different Fusion Strategy

To find the best fusion strategy, we experimented with a number of different ones. In this work, three different kinds of fusion techniques were used. The purpose of these fusion solutions was to combine the encoding of the transformer encoder and the STGCN-LSTM encoder, two distinct architectural streams. Three different fusion procedures were used: Fusing with Summation, Fusing with a Linear Layer, and Fusion Using an LSTM Layer. We decided to include Fused with Summation in our technique because it outperforms the other strategies.

| Fusion Strategy | BLEU1 | BLEU2 | BLEU3 | BLEU4 | reduced BLEU | Dataset Split |
|---|---|---|---|---|---|---|
| LSTM layer | 38.33 | 27.62 | 21.11 | 16.96 | 8.51 | Validation |
| | 37.96 | 28.05 | 21.72 | 17.66 | 6.59 | Test |
| Linear layer | 40.13 | 29.04 | 22.25 | 17.9 | 8.29 | Validation |
| | 40.5 | 29.54 | 22.67 | 18.34 | 5.97 | Test |
| Summation | 40.64 | 29.8 | 23.27 | 19.01 | 8.76 | Validation |
| | 40.28 | 29.32 | 22.48 | 18.02 | 6.57 | Test |

## 5.6 System Demonstration

This section demonstrates the functionality of our system across the four languages we actively engage with. In the system, the initial step involves selecting the language for translation. The available options include BdSL, ASL, CSL and DGH. Following the language selection, the user chooses a recorded video file. Subsequently, the system initiates the transcription process. During this phase, the system plays the sign language video and displays the reference and prediction in a text box. For German and Chinese texts, an English translation is also provided to assess the accuracy of the translated sentences.

### 5.6.1 Translation of DGS

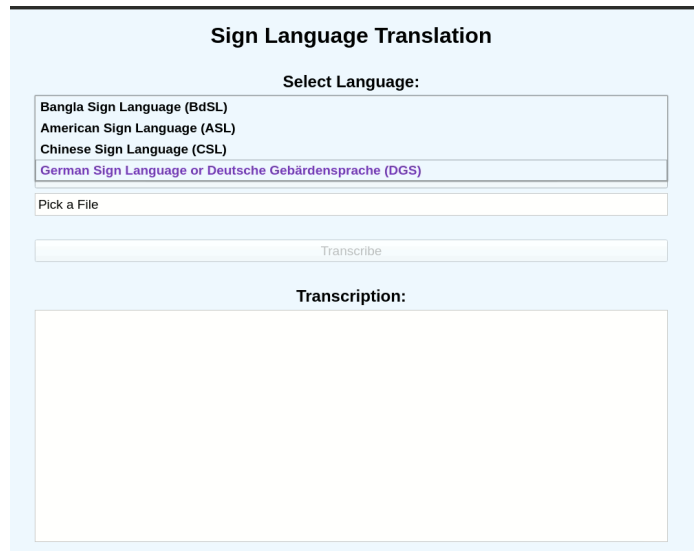

Figure 5.1: German Sign Language or deutsche Gebärdensprache(DGS) is selected as the language to translate from

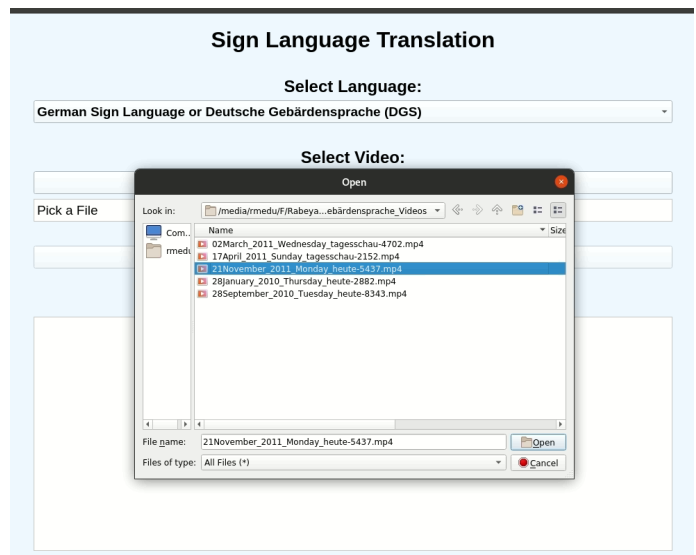

Figure 5.2: The recorded video file of a DGH is selected to translate

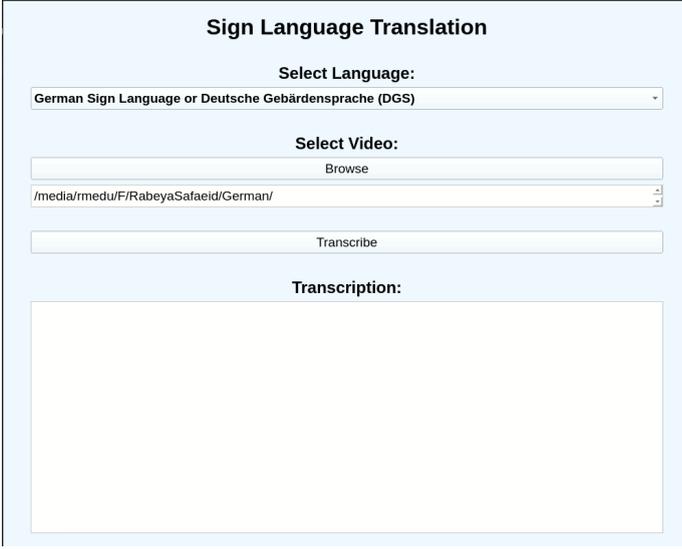

Figure 5.3: Initiate the translation task by clicking the Transcribe button

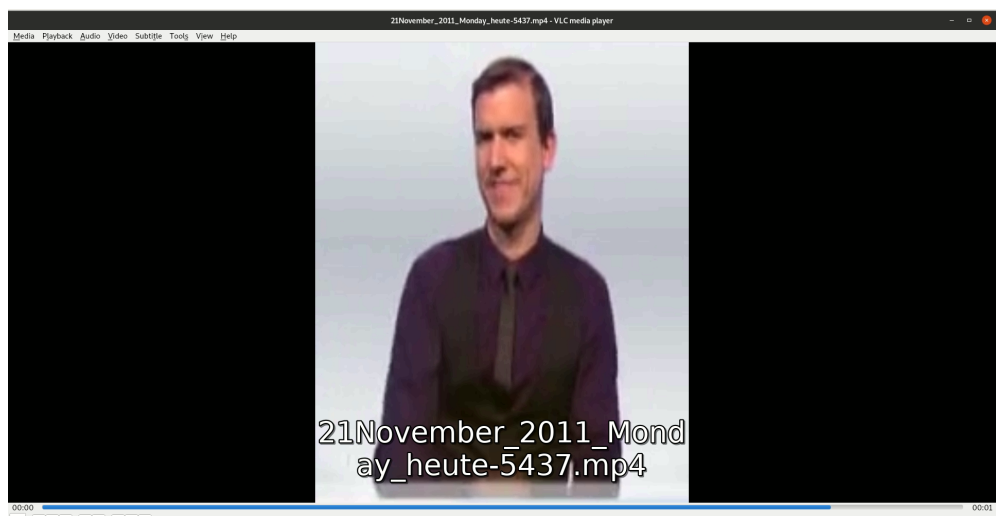

Figure 5.4: Plays the selected DGH video file

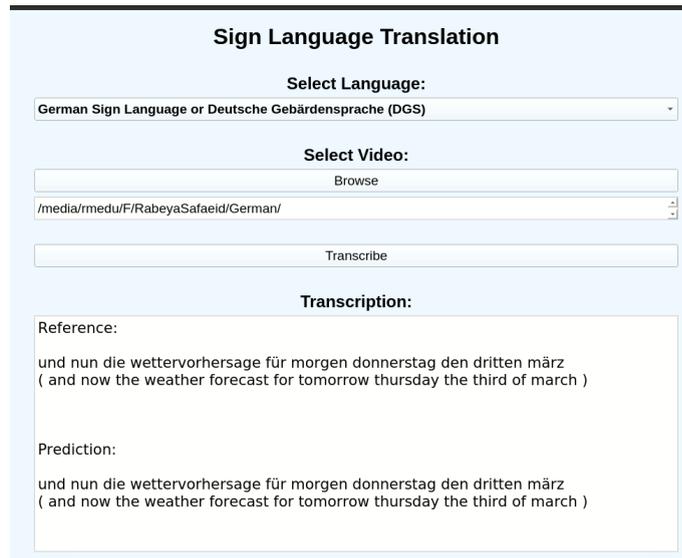

Figure 5.5: Displays the reference and predicted translated text in German and English

## 5.6.2  Translation of CSL

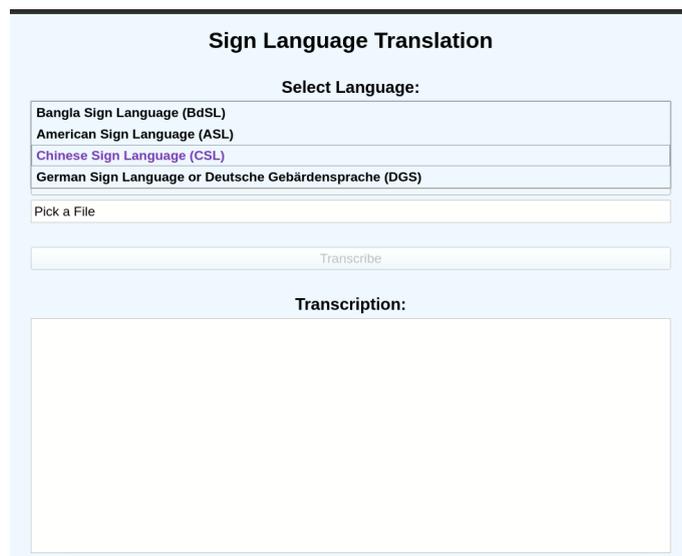

Figure 5.6: Chinese Sign Language (CSL) is selected as the language to translate from

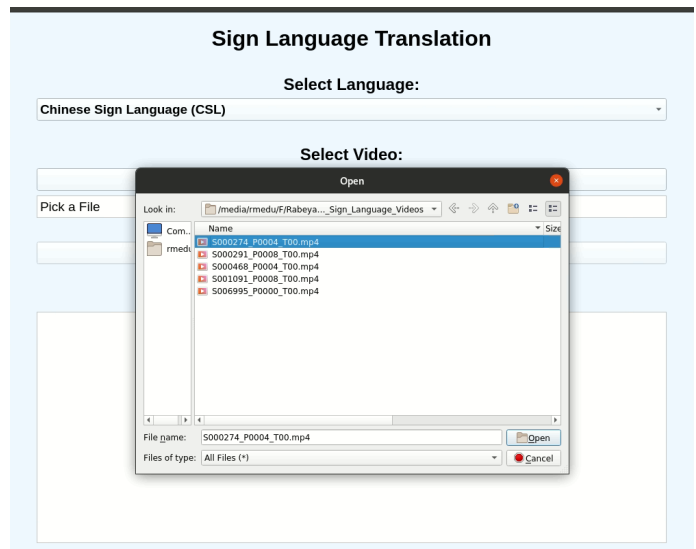

Figure 5.7: The recorded video file of a CSL is selected to translate

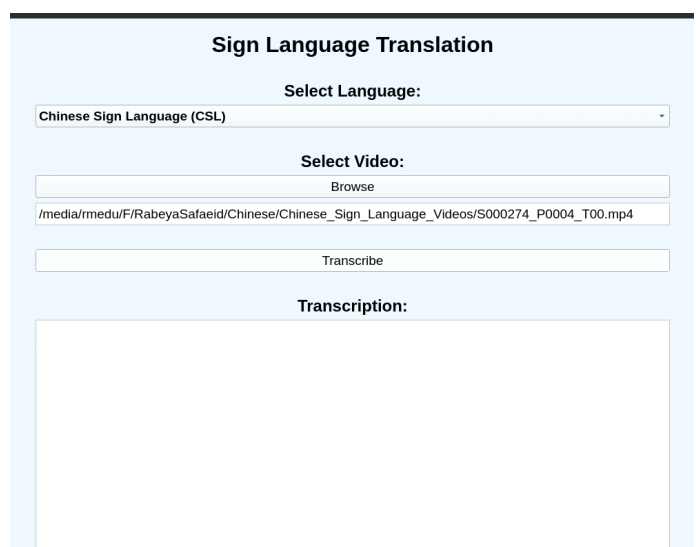

Figure 5.8: Initiate the translation task by clicking the Transcribe button

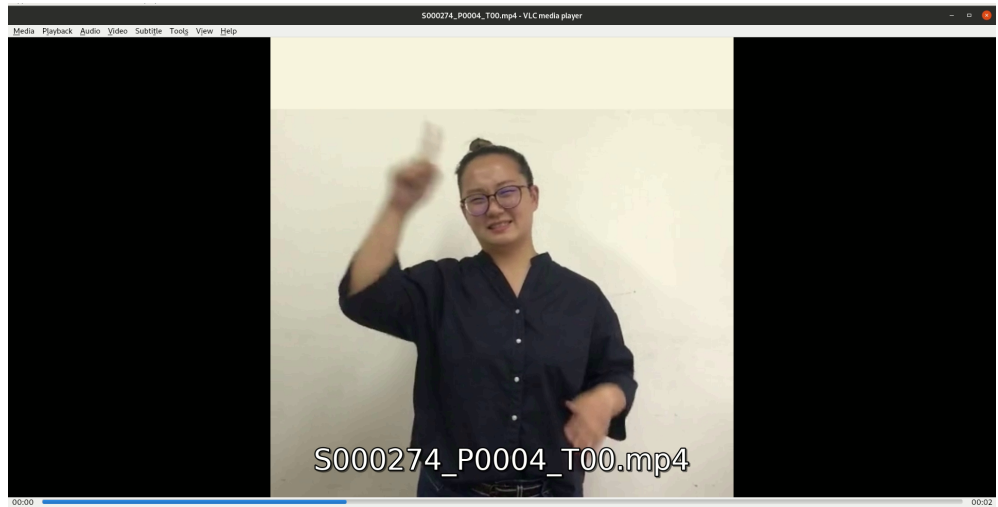

Figure 5.9: Plays the selected CSL video file

**Sign Language Translation**

**Select Language:**

Chinese Sign Language (CSL)

**Select Video:**

Browse

/media/rmedu/F/RabeyaSafaeid/Chinese/Chinese_Sign_Language_Videos/S000274_P0004_T00.mp4

Transcribe

**Transcription:**

Reference:

今天天气有点冷。
( It's a bit cold today. )

Prediction:

今天的天气很冷。
( It's cold today. )

Figure 5.10: Displays the reference and predicted translated text in Chinese and English

### 5.6.3 Translation of ASL

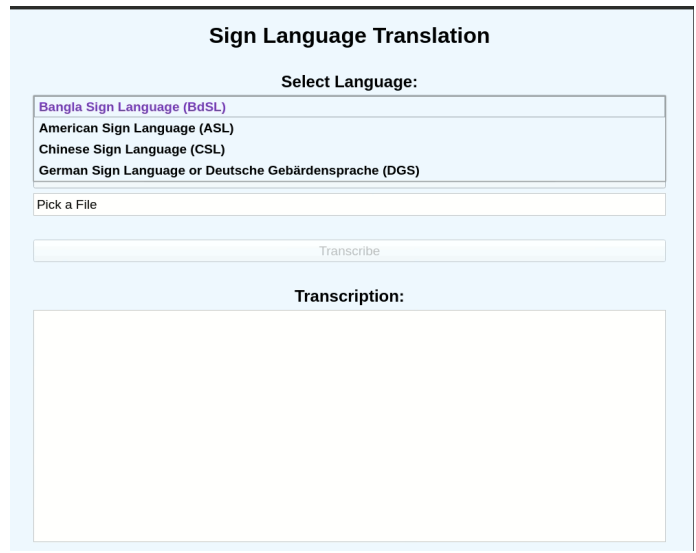

Figure 5.11: American Sign Language (ASL) is selected as the language to translate from

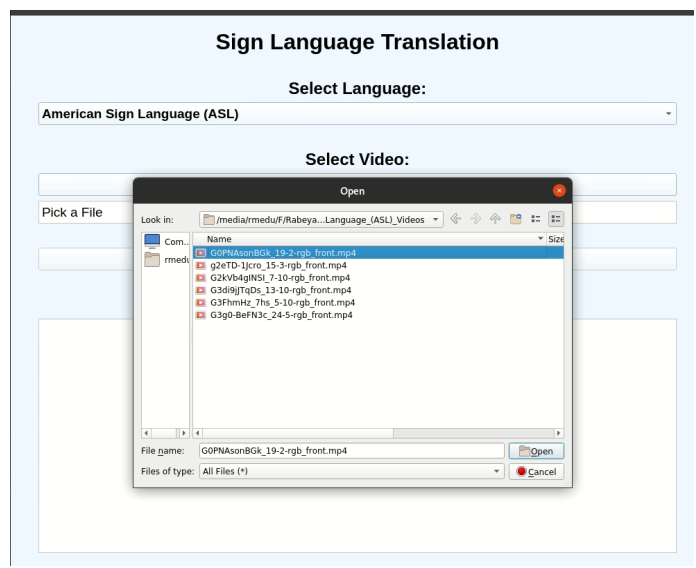

Figure 5.12: The recorded video file of a ASL is selected to translate

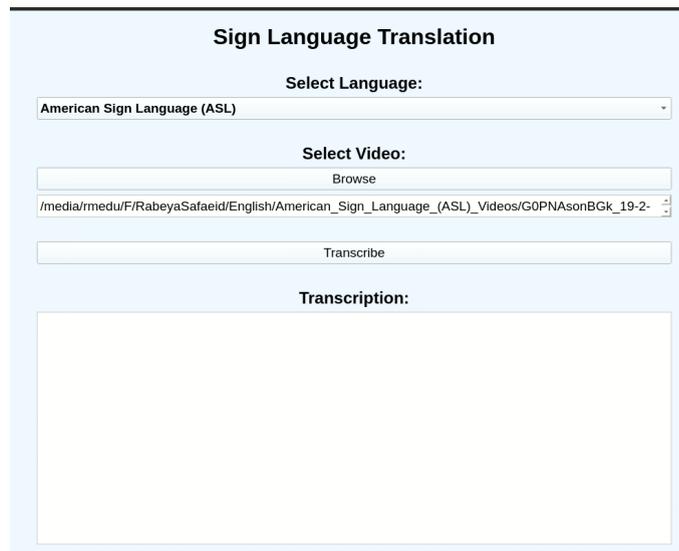

Figure 5.13: Initiate the translation task by clicking the Transcribe button

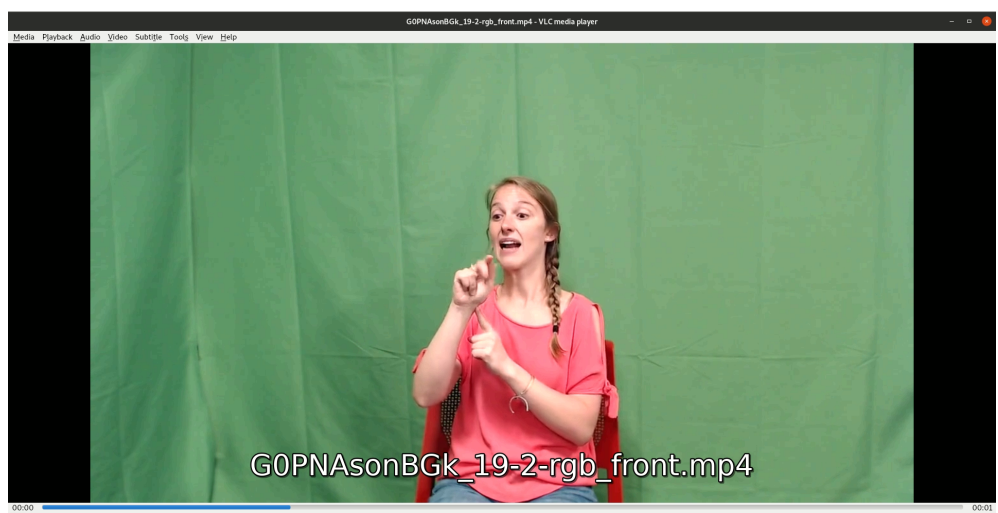

Figure 5.14: Plays the selected ASL video file

Figure 5.15: Displays the reference and predicted translated text in English

### 5.6.4   Translation of BdSL

Figure 5.16: Bangla Sign Language (BdSL) is selected as the language to translate from

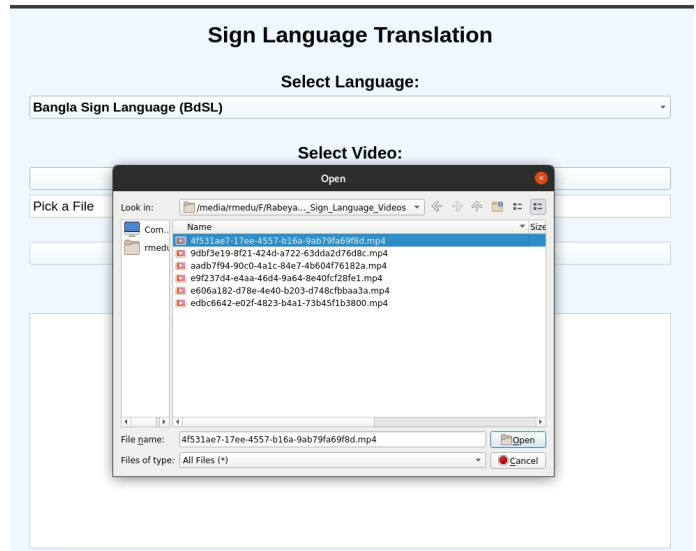

Figure 5.17: The recorded video file of a BdSL is selected to translate

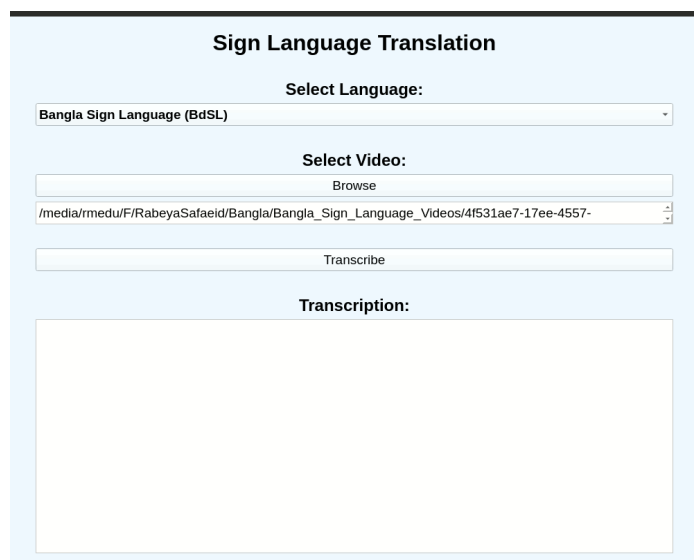

Figure 5.18: Initiate the translation task by clicking the Transcribe button

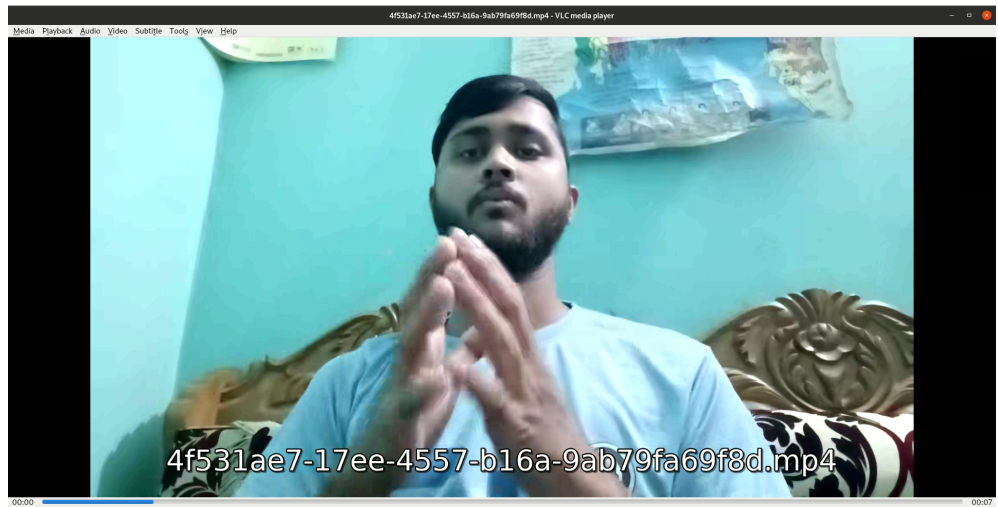

Figure 5.19: Plays the selected BdSL video file

**Sign Language Translation**

**Select Language:**

Bangla Sign Language (BdSL)

**Select Video:**

Browse

/media/rmedu/F/RabeyaSafaeid/Bangla/Bangla_Sign_Language_Videos/4f531ae7-17ee-4557-

Transcribe

**Transcription:**

Reference:

অনেক কাজ করতে হবে।

Prediction:

অনেক কাজ করতে হবে।

Figure 5.20: Displays the reference and predicted translated text in Bangla

# Chapter 6

# Conclusion

The communication challenges encountered by individuals with hearing impairments give rise to adverse emotional states, including isolation, frustration, and psychological concerns. This not only affects personal well-being but also results in substantial economic implications, with an estimated annual healthcare cost of USD 750 billion[15]. The developed methodology in this research holds the potential to contribute to the creation of automated systems for sign language interpretation. The successful implementation of such technologies can significantly enhance inclusivity and accessibility for the entire community. Notably, the deaf and hard-of-hearing population in Bangladesh stands to gain substantial benefits from our endeavors to establish an end-to-end solution for recognizing Bangla sign language. The significance of our research lies in its contribution to addressing the communication barriers faced by the deaf and hard-of-hearing community. The successful development of an automated sign language interpretation system serves as a foundation for fostering inclusivity and accessibility.

Our research delved into four diverse sign language datasets, encompassing ASL, DGH, CSL, and BDSL. Each dataset presented unique challenges, with the Bangla Sign Language dataset posing a distinctive hurdle due to its dynamic background, unlike the others with a plain backdrop or a green screen. Despite these challenges, our efforts yielded good results across all datasets, culminating in a robust demonstration of the system's capabilities.

Moving forward, our primary goal is to strengthen the system's resilience, facilitating the real-time processing of video data beyond the constraints of solely relying on pre-recorded inputs. This future trajectory aims to augment the practical utility of the system in real-world scenarios, rendering it more responsive and adaptable to diverse communication situations. There is still ample room for improvement in SLT. Presently, the best translation scores fall considerably short of declaring the task as solved. All existing models, including ours, demand substantial computational resources. In our future plans, we intend to develop a more lightweight version to enable its operation on mobile devices with lower resource requirements.

In conclusion, our research not only addresses the immediate challenges faced by the deaf and hard-of-hearing community but also sets the stage for continuous improvement and innovation in the field of sign language interpretation. By striving for real-time processing, increased accessibility, and reduced computational demands, we aim to make a lasting impact on the lives of those with hearing impairments, fostering a more inclusive and connected world.

# Appendix A: List of Acronyms

BdSL  Bangla Sign Language

DGS  Deutsche Gebärdensprache [German Sign Language]

CSL  Chinese Sign Language

ASL  American Sign Language

BLUE-n  Bilingual Evaluation Understudy - n-gram

ROUGE-L  Recall-Oriented Understudy for Gisting Evaluation - Longest Common
        Subsequence

WHO  World Health Organization

PCA  Principal Component Analysis

CNN  Convolutional Neural Network

GCN  Graph Convolution Network

TCN  Temporal Convolution Network

STGNN  Spatio-Temporal Graph Convolution Network

LSTM  Long Short-Term Memory

ConvLSTM  Convolutional Long Short-Term Memory

RNN  Recurrent Neural Network

GNN  Graph Neural Networks

ReLU  Rectified Linear Unit